\documentclass{article}

\usepackage{graphicx}
\usepackage{amsmath,amssymb} 
\usepackage{color}
\usepackage{url}
\usepackage{hyperref}
\usepackage{times}
\usepackage{natbib}

\usepackage{algorithm}
\usepackage{algorithmic}
\usepackage[accepted]{icml2014}

\icmltitlerunning{HDR by Perceptual Merging}

\begin{document}

\twocolumn[ \icmltitle{High Dynamic Range Imaging \\ by Perceptual Logarithmic Exposure Merging}

\icmlauthor{Corneliu Florea}{corneliu.florea@upb.ro} \icmladdress{Image Processing and Analysis Laboratory\\
                University "Politehnica" of Bucharest, Romania,
                Address Splaiul Independen\c{t}ei 313}
\icmlauthor{Constantin Vertan}{constantin.vertan@upb.ro} \icmladdress{Image Processing and Analysis Laboratory\\
                University "Politehnica" of Bucharest, Romania,
                Address Splaiul Independen\c{t}ei 313}
\icmlauthor{Laura Florea}{laura.florea@upb.ro} \icmladdress{Image Processing and Analysis Laboratory\\
                University "Politehnica" of Bucharest, Romania,
                Address Splaiul Independen\c{t}ei 313}

\icmlkeywords{ Logarithmic Image Processing, Human Visual System,  High Dynamic Range}

\vskip 0.3in

]

\begin{abstract}
   In this paper we emphasize a similarity between the Logarithmic-Type Image Processing (LTIP) model
    and the Naka-Rushton model of the Human Visual System (HVS). LTIP is a derivation of the Logarithmic
    Image Processing (LIP), which further replaces the logarithmic function with a ratio of polynomial
    functions. Based on this similarity, we show that it is possible to present an unifying framework
    for the High Dynamic Range (HDR) imaging problem, namely that performing exposure merging under the
    LTIP model is equivalent to standard irradiance map fusion. The resulting HDR algorithm is shown to
    provide high quality in both subjective and objective evaluations.
\end{abstract}


\section{Introduction}
Motivated by the limitation of digital cameras in capturing real scenes with large lightness
dynamic range, a category of image acquisition and processing techniques, collectively named High
Dynamic Range (HDR) Imaging, gained popularity. To acquire HDR scenes, consecutive frames with
different exposures are typically acquired and combined into a HDR image that is viewable on
regular displays and printers.

In parallel, Logarithmic Image Processing (LIP) models were introduced as an alternative to image
processing with real based operations. While initially modelled from the cascade of two
transmitting filters \cite{Jour:1987}, later it was shown that the LIP models can be generated by
the homomorphic theory and they have a cone space structure \cite{Deng:1995}. The initial model was
shown to be compatible with the Weber-Fechner perception law \cite{Pinoli:2007}, which is not
unanimously accepted \cite{Stevens:1961}. Currently, most global Human Visual System (HVS) models
are extracted from the Naka-Rushton equation of photoreceptor absorption of incident energy and are
followed by further modelling of the local adaptation. We will show in this paper that the new LIP
extension model introduced in \cite{Vertan:2008} is consistent with the global human perception as
described by the Naka-Rushton model. The model  no longer uses a logarithmic generative function
but only a logarithmic-like function, hence it will be named Logarithmic Type Image Processing
(LTIP) model. In such a case, the generative function of the LTIP model transfers the radiometric
energy domain into human eye compatible image domain; thus it mimics, by itself and by its inverse,
both the camera response function and the human eye lightness perception.

The current paper claims three contributions. Firstly, we show that the previously introduced LTIP
model is compatible with Naka-Rushton/Michaelis-Menten model of the eye global perception.
Secondly, based on the previous finding, we show that it is possible to treat two contrasting HDR
approaches unitary if the LTIP model framework is assumed. Thirdly, the reinterpretation of the
exposure merging algorithm \cite{Mertens:2007} under the LTIP model produces a new algorithm that
leads to qualitative results.

The paper is constructed as follows: in Section \ref{Sect:HDR_state} we present a short overview of
the existing HDR trends and we emphasize their correspondence with human perception. In Section
\ref{Sect:FLIP}, state of the art results in the LIP framework and the usage of the newly
introduced LTIP framework for the generation of models compatible with human perception are
discussed. In Section \ref{Sect:FLIPbasedTMO} we derive and motivate the submitted HDR imaging,
such that in Section \ref{Sect:Implement} we discuss implementation details and achieved results,
ending the paper with discussion and conclusions.

\section{Related work}
\label{Sect:HDR_state}

\begin{table*}
\centering \caption{Comparison of the two main approaches to the HDR problem.
\label{Tab:HDR_Comparison}}
    \begin{tabular}{|c|c|c|c|c|}
    \hline
        \textbf{Method} & \textbf{CRF recovery} & \textbf{Fused Components} & \textbf{Fusion Method} &
        \textbf{Perceptual} \\ \hline
            \begin{tabular}{c} Irradiance fusion \\ \cite{Debevec:1997} \end{tabular} & Yes & Irradiance Maps &
            \begin{tabular}{c} Weighted \\ convex combination \end{tabular} & Yes (via TMO)
        \\ \hline
            \begin{tabular}{c} Exposure fusion \\ \cite{Mertens:2007} \end{tabular} & No & Acquired Frames &
            \begin{tabular}{c} Weighted \\ convex combination \end{tabular} & No \\
    \hline
    \end{tabular}
\end{table*}

\begin{figure*}[tb]
 \centering
    \begin{tabular}{cc}

   \includegraphics[width=0.35\linewidth]{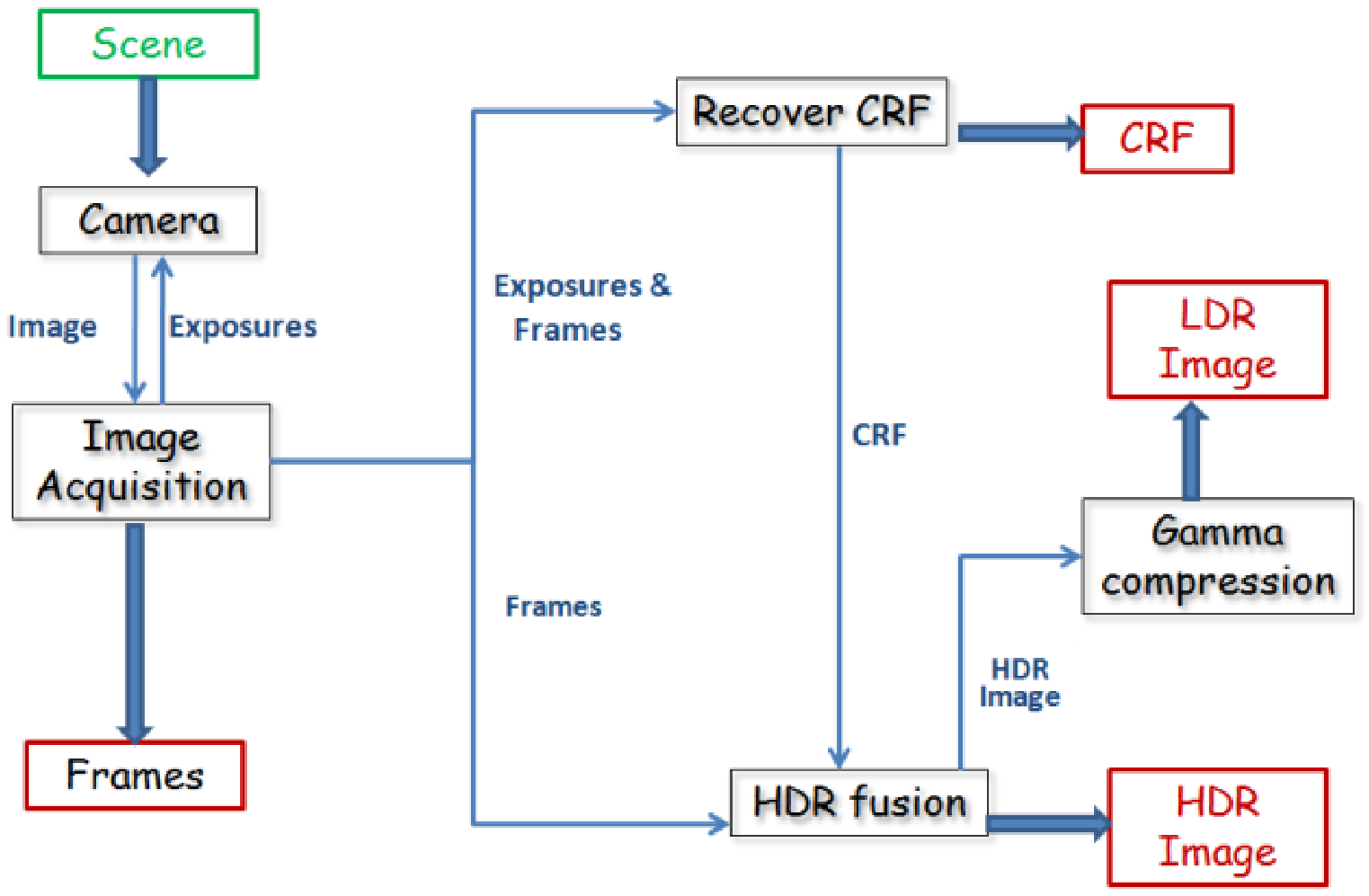} &
   \includegraphics[width=0.35\linewidth]{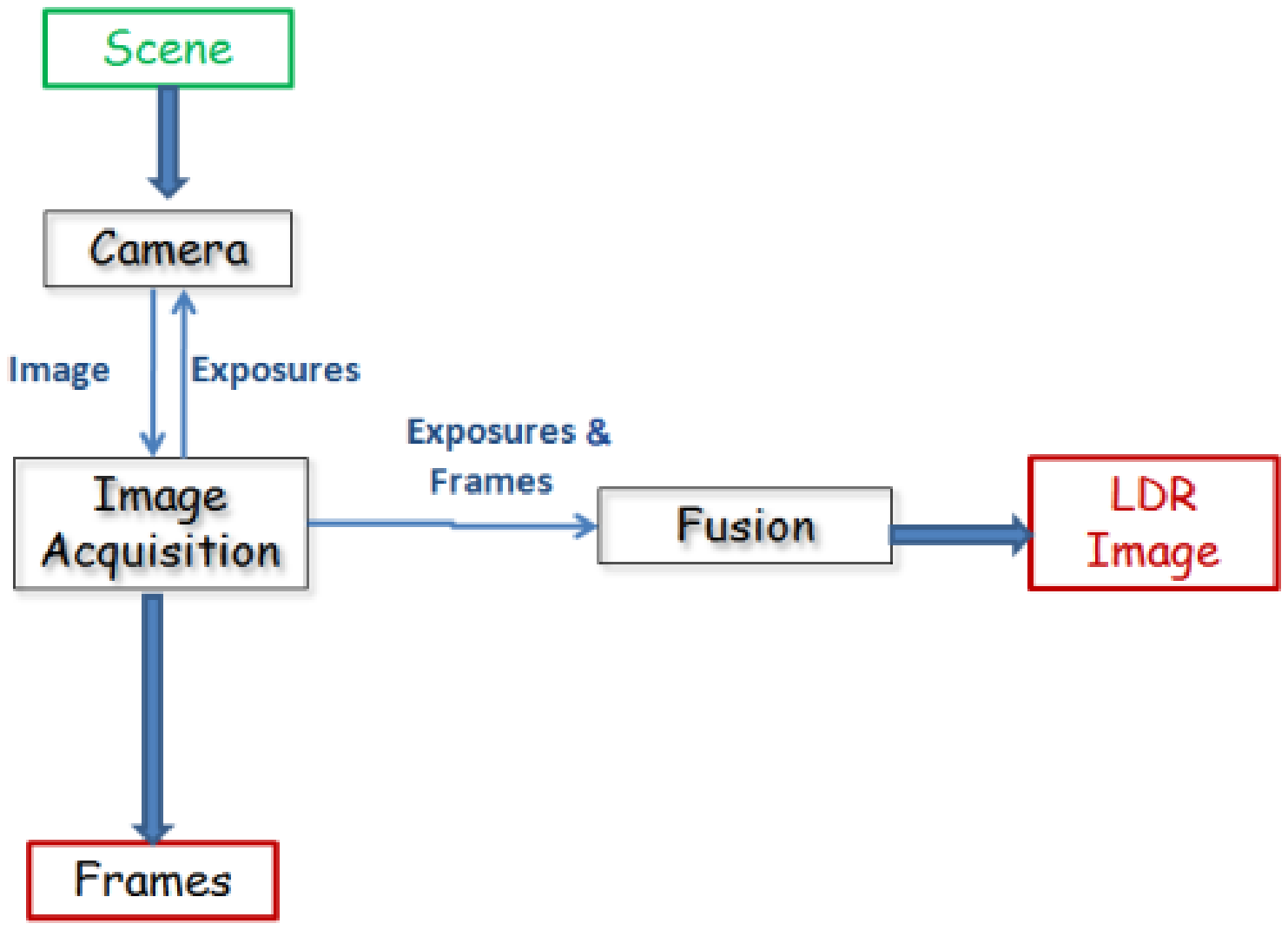} \\
   (a) & (b)
    \end{tabular}
   \caption{  HDR imaging techniques: (a) irradiance maps fusion as described in Debevec and Malik \yrcite{Debevec:1997}
   and (b) exposure fusion as described in Mertens et al.\yrcite{Mertens:2007}.
   Irradiance map fusion relies on
   inverting the Camera Response Function (CRF) in order to return to the irradiance domain, while the exposure
   fusion works directly in the image domain, and thus avoiding the CRF reversal.  }
     \label{Fig:HDRSchema}
\end{figure*}

\begin{table*}[tb]
    \centering
    \caption{The classical LIP model introduced by Jourlin and Pinolli,
    the logarithmic type (LTIP) model 
    with the  basic operations and the parametric extension of the LTIP model. $D$ is
    the upper bound of the image definition set (typically $D=255$ for \texttt{unsigned int} representation
    or $D=1$ for \texttt{float} image representation).
     \label{Tab:LTIPS}
     }
    \begin{tabular}{|c|c|c|c|c|}
        \hline Model & Domain & Isomorphism &
                \begin{tabular}{c} Addition \\ $u\oplus v$ \end{tabular} &
                \begin{tabular}{c} Scalar\\multiplication \\ $\alpha \otimes u$ \end{tabular} \\
        \hline
          LIP &
           $\mathcal{D_{\phi}} = (-\infty ; D]$ &
           $\Phi(x)=-D\log \frac{D}{D-x}$   &
           $ u + v + \frac{u v}{D}$ &
           $ D - D\left (1-\frac{u}{D}\right ) ^\alpha$\\
        \hline
         LTIP &
            $\mathcal{D_{\phi}} = [0 ; 1)$ &
            $\Phi(x)=\frac{x}{1-x}$ &
            $ 1-\frac{(1-u)( 1- v)}{1 -uv}$ &
            $\frac{\alpha u}{1+(\alpha-1)u}$ \\
         \hline
        \begin{tabular}{c} Parametric \\ LTIP \end{tabular} &
            $\mathcal{D_{\phi}} = [0 ; 1)$ &
            $\Phi_m(x)=\frac{x^m}{1-x^m}$ &
            $\sqrt[m]{1-\frac{(1-u^m)(1-v^m)}{1-u^mv^m}}$ &
            $ u \sqrt[m]{\frac{\alpha}{1+(\alpha-1)u^m}}$ \\
         \hline
    \end{tabular}
\end{table*}

The typical acquisition of a \textbf{High Dynamic Range} image relies on the ``Wyckoff Principle'',
that is differently exposed images of the same scene capture different information due to the
differences in exposure \cite{Mann:1995}. Bracketing techniques are used in practice to acquire
pictures of the same subject but with consecutive exposure values. These pictures are, then, fused
to create the HDR image.

For the fusion step two directions are envisaged. The first direction, named \emph{irradiance
fusion}, acknowledges that the camera recorded frames are non-linearly related to the scene
reflectance and, thus, it relies on the irradiance maps retrieval from the acquired frames, by
inverting the camera response function (CRF), followed by fusion in the irradiance domain. The
fused irradiance map is compressed via a tone mapping operator (TMO) into a displayable low dynamic
range (LDR) image. The second direction, called \emph{exposure fusion}, aims at simplicity and
directly combines the acquired frames into the final image. A short comparison between these two is
presented in Table \ref{Tab:HDR_Comparison} and detailed in further paragraphs.

\subsection{Irradiance fusion}

Originating in the work of Debevec and Malik \yrcite{Debevec:1997}, the schematic of the irradiance
fusion may be followed in Fig. \ref{Fig:HDRSchema} (a). Many approaches were schemed for
determining the CRF \cite{Grossberg:2004}. We note that the dominant shape is that of a gamma
function \cite{Mann:2001}, trait required by the compatibility with the HVS.

After reverting the CRF, the irradiance maps are combined, typically by a convex combination
\cite{Debevec:1997}, \cite{Robertson:1999}. For proper displaying, a tone mapping operator (TMO) is
then applied on the HDR irradiance map to ensure that in the compression process all image details
are preserved. For this last step, following Ward's proposal  \cite{Ward:97}, typical approaches
adopt a HVS-inspired function for domain compression, followed by local contrast enhancement. For a
survey of the TMOs we refer to the paper of Ferradans et al. \yrcite{Ferradans:2011} and to the
book by Banterle et al.\yrcite{Banterle:2011}.

Among other TMO attempts, a notable one was proposed by Reinhard et al. \yrcite{Reinhard:2002}
which, inspired by Ansel Adams' Zone System, firstly applied a logarithmic scaling to mimic the
exposure setting of the camera, followed by dodging-and-burning (selectively and artificially
increase and decrease image values for better contrast) for the actual compression. Durand et al.
\yrcite{Durand:2002} separated, by means of a bilateral filter, the HDR irradiance map into a base
layer that encoded large scale variations (thus, needing range compression) and into a detail
preserving layer to form an approximation of the image pyramid. Fattal et al. \yrcite{Fattal:2002}
attenuated the magnitude of large gradients based on a Poisson equation. Drago et al.
\yrcite{Drago:2003} implemented a logarithmic compression of luminance values that matches the HVS.
Krawczyk et al. \yrcite{Krawczyk:2005} implemented the Gestalt based anchoring theory  of Gilchrist
Gilchrist et al. \yrcite{Gilchrist:1999} to divide the image in frameworks and performed range
compression by ensuring that frameworks are well-preserved. Banterle et al. \yrcite{Banterle:2012}
segmented the image into luminance components and applied independently the TMOs introduced in
Drago et al. \cite{Drago:2003} and in Reinhard et al. \cite{Reinhard:2005} for further adaptive
fusion based on previously found areas. Ferradans et al. \yrcite{Ferradans:2011} proposed an
elaborated model of the global HVS response and pursued local adaptation with an iterative
variational algorithm.

Yet, as irradiance maps are altered with respect to the reality by the camera optical systems,
additional constraints are required for a perfect match with the HVS. Hence, this category of
methods, while being theoretically closer to the pure perceptual approach, requires supplementary
and costly constraints and significant computational resources for the CRF estimation and for the
TMO implementation.

\subsection{Exposure merging}

Noting the high computational cost of the irradiance maps fusion, Mertens et al.
\yrcite{Mertens:2007} proposed to implement the fusion directly in the image domain; this approach
is described in Fig. \ref{Fig:HDRSchema} (b). The method was further improved for robustness to
ghosting artifacts and details preservation in HDR composition by Pece and Kautz
\yrcite{Pece:2010}.

Other developments addressed the method of computing local contrast to preserve edges and local
high dynamic range. Another expansion has been introduced by Zhang et al. \yrcite{Zhang:2012}, who
used the direction of gradient in a partial derivatives type of framework and two local quality
measures to achieve local optimality in the fusion process. Bruce \yrcite{Bruce:2014} replaced the
contrast computed by Mertens et al. \yrcite{Mertens:2007} onto a Laplacian pyramid with the entropy
calculated in a flat circular neighborhood for deducing weights that maximize the local contrast.

The exposure fusion method is the inspiration source for many commercial applications. Yet, in such
cases, exposure fusion is followed by further processing that increases the visual impact of the
final image. The post--processing includes contrasting, dodging-and-burning, edge sharpening, all
merged and tuned in order to produce a surreal/fantasy-like aspect of the final image.

While being sensibly faster, the exposure fusion is not physically motivated, nor perceptually
inspired. However, while the academic world tends to favor perceptual approaches as they lead to
images that are correct from a perceptual point of view, the consumer world naturally tends to
favor images that are photographically more spectacular and the exposure merging solution pursuits
this direction.


\section{Logarithmic Type Image Processing} \label{Sect:FLIP}

Typically, image processing operations are performed using real-based algebra, which proves its
limitations under specific circumstances, like upper range overflow. To deal with such situations,
non-linear techniques have been developed \cite{Rafajlowicz:2008}. Such examples are the LIP
models. The first LIP model was constructed by Jourlin and Pinoli \cite{Jour:1987} starting from
the equation of light passing through transmitting filters.

The LIP model was further developed into a robust mathematical structure, namely a cone/vector
space. Subsequently many practical applications have been presented and an extensive review of
advances and applications for the classical LIP model is presented in \cite{Pinoli:2007}. In
parallel, other logarithmic models and logarithmic-like models were reported. In this particular
work we are mainly interested in the logarithmic-like model introduced by Vertan et al.
\yrcite{Vertan:2008}, which has a cone space structure and is named LTIP model. A summary of
existing models may be followed in \cite{Navarro:2013}. Recently parametric extensions of the LTIP
models were also introduced \cite{Panetta:2011}, \cite{Florea:2013}. The LTIP models are summarized
in Table \ref{Tab:LTIPS}.

\subsection{Relation between LIP models and HVS}
From its introduction in the 80s, the original LIP model had a strong argument being similar with
the Weber-Fechner law of contrast perception. This similarity was thoroughly discussed in
\cite{Pinoli:2007}, where it was shown that logarithmic subtraction models the increment of
sensation caused by the increment  of light with the quantity existing in the subtraction. Yet the
logarithmic model of the global perceived luminance contrast system assumed by the Weber-Fechner
model was vigorously challenged \cite{Stevens:1961} and arguments hinted to the power-law rules
\cite{Stevens:1963}. Thus, we note that the Stevens model is more inline with the LTIP model. On
the other hand, Stevens experiments were also questioned \cite{Macmillan:2005}, so it does not seem
to be a definite answer in this regard.

Still, lately, the evidence seems to favor the Naka-Rushton/Michaelis-Menten model of retinal
adaptation \cite{Ferradans:2011}, thus an important class of TMO techniques following this model
for the global adaptation step. The Naka-Rushton equation is a particular case of the
Michaelis-Menten model that expresses the hyperbolic relationship between the initial velocity and
the substrate concentration in a number of enzyme-catalyzed reactions. Such a process is the change
of the electric potential of a photoreceptor (e.g. the eye cones) membrane, $r(\mathcal{I})$ due to
the absorption of light of intensity $\mathcal{I}$. The generic form, called Michaelis-Menten
equation \cite{Valeton:1983} is:
\begin{equation}
    r(\mathcal{I}) = \frac{\Delta V(\mathcal{I})}{\Delta V_{\max}} = \frac{\mathcal{I}^n}{\mathcal{I}^n + \mathcal{I}_S^n}
   \label{Eq:MichMenten}
\end{equation}
where $\Delta V_{\max}$ is the maximum difference of potential that can be generated,
$\mathcal{I}_S^n$ is  the light level at which the photoreceptor response is half maximal
(semisaturation level) and $n$ is a constant. Valeton and van Norren \yrcite{Valeton:1983}
determined that $n=0.74$ for rhesus monkey. If $n=1$ the Naka-Rushton equation \cite{Naka:1966} is
retrieved as a particular case of the Michaelis-Menten model:
\begin{equation}
    r(\mathcal{I}) = \frac{\Delta V(\mathcal{I})}{\Delta V_{\max}} = \frac{\mathcal{I}}{\mathcal{I} + \mathcal{I}_S}
    \label{Eq:NakaRush}
\end{equation}

For the TMO application, it is considered that the electric voltage in the right is a good
approximation of the perceived brightness \cite{Ferradans:2011}. Also, it is not uncommon
\cite{Meylan:2007} to depart from the initial meaning of semisaturation for $\mathcal{I}_S$ (the
average light reaching the light field) and to replace it with a convenient chosen constant. TMOs
that aimed to mimic the Naka-Rushton model \cite{Reinhard:2002}, \cite{Tamburino:2008} assumed that
the HDR map input was $\mathcal{I}$ and obtained the output as $r(\mathcal{I})$.

On the other hand, the generative function of the LIP model maps the image domain onto the real
number set. The inverse function acts as a homomorphism between the real number set and the closed
space that defines the domain of LIP. For the LTIP model, the generative function is
$\Phi_V(x)=\frac{x}{1-x}$ while the inverse is:
\begin{equation}
     \Phi_V^{-1}(y)=\frac{y}{y+1}
     \label{Eq:LTIP_genInv}
\end{equation}

The inverse function (Eq. (\ref{Eq:LTIP_genInv}) ) mimics the Naka-Rushton  model --Eq.
(\ref{Eq:NakaRush}), with the difference that instead of the semi-saturation, $\mathcal{I}_S$, as
in the original model, it uses full saturation. Given this observation, we interpreted
logarithmic-like model as the mapping of the irradiance intensity (which are defined over the real
number set) onto photoreceptor acquired intensities, i.e human observable chromatic intensity.
While the logarithmic-like model is only similar and not identical with the Naka-Rushton model of
the human eye, it has the strong advantage of creating a rigorous mathematical framework of a cone
space.


\subsection{Relation between the LIP models and CRF}
The dominant non-linear transformation in the camera pipe-line is the gamma adjustment necessary to
adapt the image to the non-linearity of the display and respectively of the human eye. The entire
pipeline is described by the Camera Response Function (CRF) which, typically, has a gamma shape
\cite{Grossberg:2004}.

\begin{figure}[t]
\centering
    \begin{tabular}{cc}
        \includegraphics[width=0.45\linewidth]{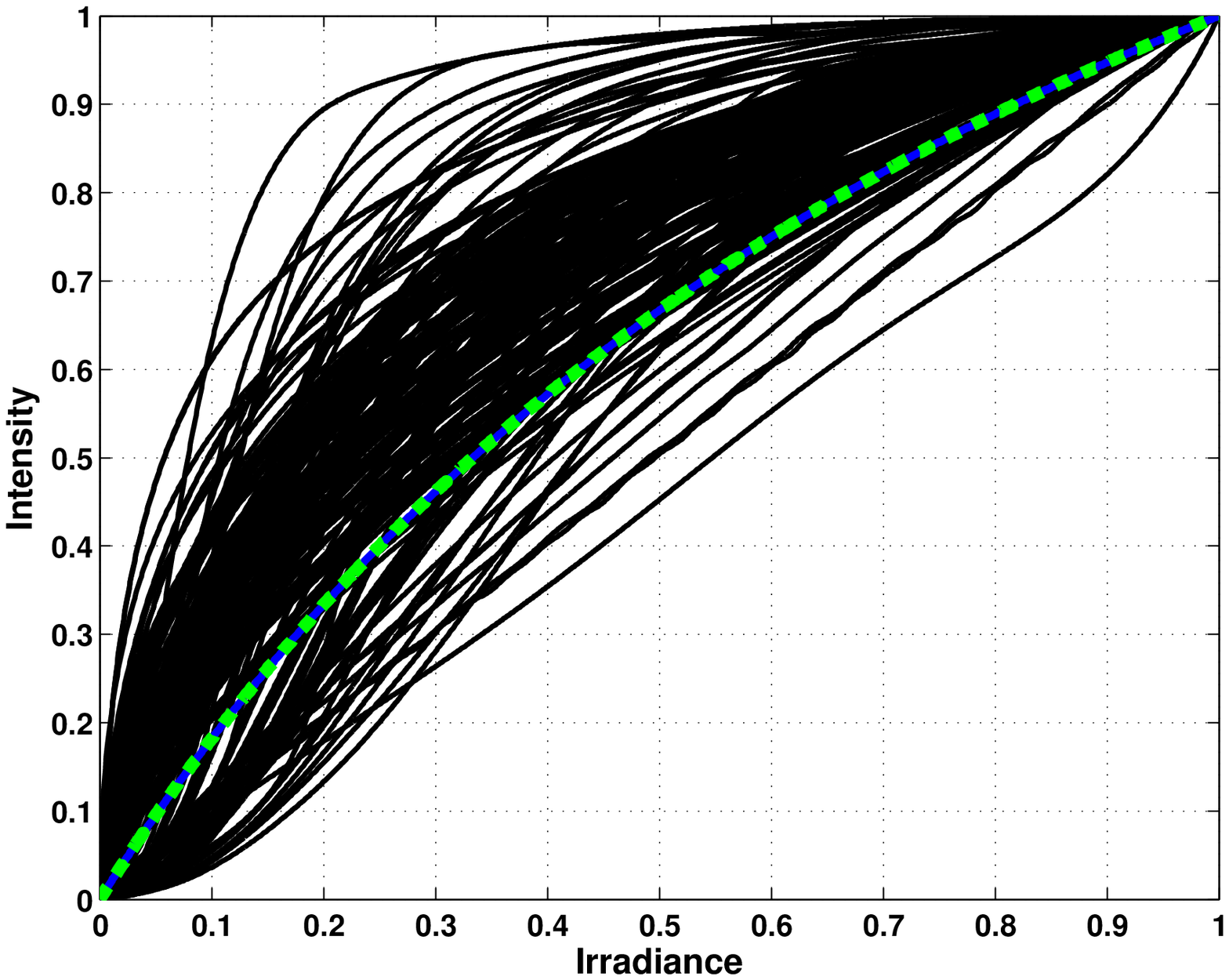} &
        \includegraphics[width=0.45\linewidth]{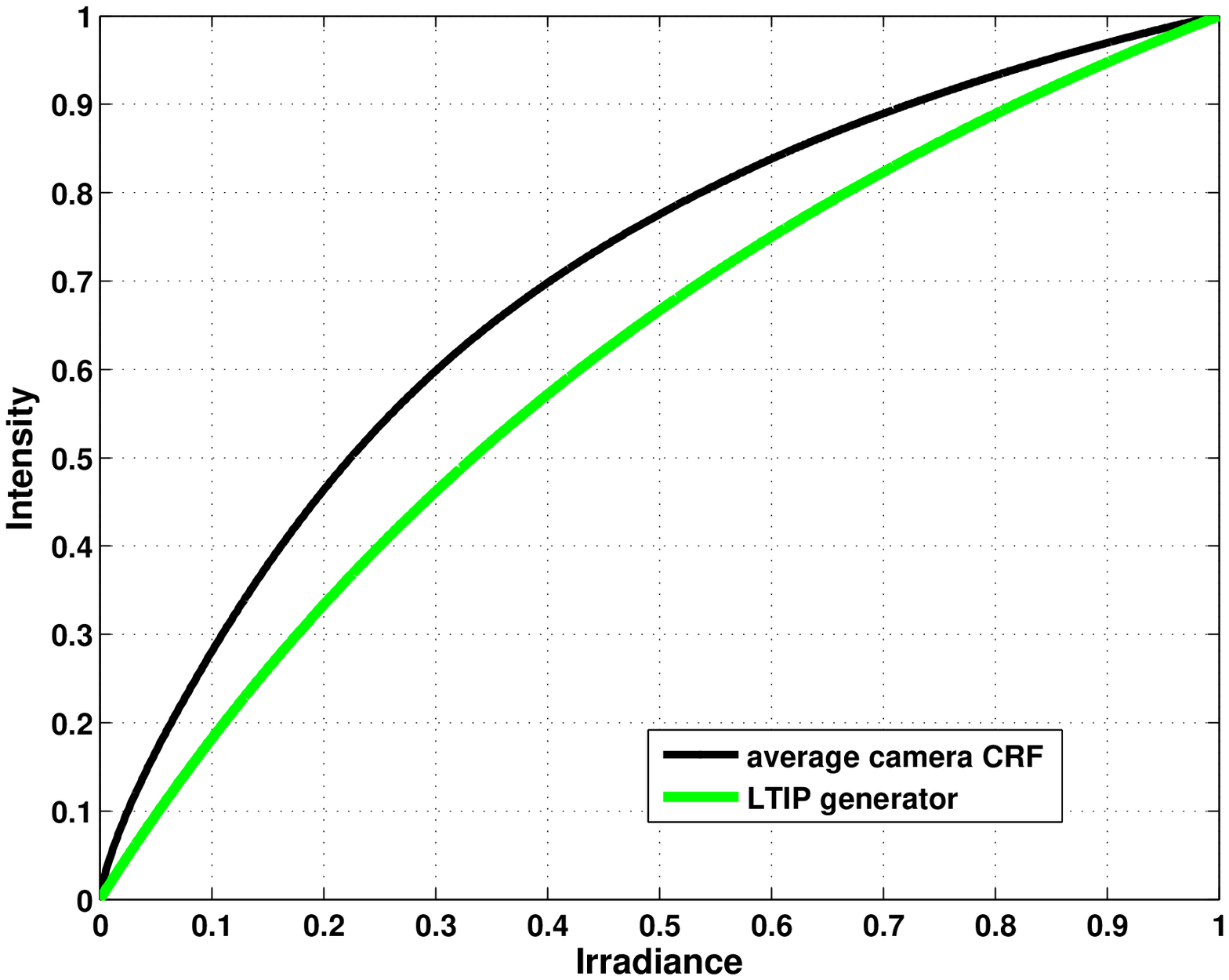} \\
        (a) & (b)
    \end{tabular}
    \caption{The relation between the LTIP generative function $\Phi_V$ and camera response functions as
    recorded in DoRF database: (a) full database and (b) average CRF (black) with respect to the LTIP function (green).}
    \label{Fig:CRFs}
\end{figure}

It was previously pointed to the similarity between the LTIP generative function and the CRF
\cite{Florea:2013}. To show the actual relation between the LTIP generative function and the CRF,
we considered the Database of Response Functions (DoRF) \cite{Grossberg:2004} which consists of 200
recorded response functions of digital still cameras and analogue photographic films. These
functions are shown in Fig. \ref{Fig:CRFs} (a); to emphasize the relation, in subplot (b) of the
same figure we represented only the LTIP generative function and the average CRF. As one may see,
while the LTIP generative function is not identical to the average CRF, there do exist camera and
films that have a response function identical to the LTIP generative function.

To improve the contrast and the overall appearance of the image, some camera models add an S-shaped
tone mapping that no longer follows the Naka-Rushton model. In such a case, a symmetrical LTIP
model, such as the one described in (Navarro et al. 2013) has greater potential to lead to better
results.

\section{HDR by Perceptual Exposure Merging}
\label{Sect:FLIPbasedTMO}

Once that camera response function, $g$, has been found, the acquired images $\mathbf{f}_i$ are
turned into irradiance maps, $\mathbf{E}_i$ by \cite{Debevec:1997}:

\begin{equation}
    \mathbf{E}_i(k,l) = \frac{g^{-1}\left(\mathbf{f}_i(k,l)\right)}{\Delta t}
\end{equation}
where $\Delta t$ is the exposure time and $(k,l)$ is the pixel location. Further, the HDR
irradiance map is calculated as the weighted sum of the acquired irradiance maps
\cite{Debevec:1997}, \cite{Robertson:1999}:

\begin{equation}
    \mathbf{E}_{HDR}(k,l) = \frac{\sum_{i=1}^N w\left(\mathbf{f}_i(k,l) \right)\cdot
    \mathbf{E}_i(k,l)}{\sum_{i=1}^N w\left(\mathbf{f}_i(k,l)\right)}
    \label{Eq:Hdr_rad_Convex}
\end{equation}
where $w \left( \textbf{f}_i(k,l) \right)$ are weights depending on the chosen algorithm and $N$ is
the number of frames.

However, we stress that the weights are scalars with respect to image values. This means that their
sum is also a scalar and we denote it by:
\begin{equation}
    \mathbf{\eta} = \sum_{i=1}^N w\left(\mathbf{f}_i(k,l) \right)
\end{equation}

Taking into account that the CRF may be approximated by the LTIP generative function $g$ and, also,
that the final image was achieved by a tone mapping operator from the HDR irradiance map, we may
write that:
\begin{equation}
    \mathbf{f}_{HDR}(k,l) = g\left(\mathbf{E}_{HDR}(k,l)\right)
\end{equation}

If one expands the HDR irradiance map using Eq. (\ref{Eq:Hdr_rad_Convex}), he will obtain:
\begin{equation}
\label{Eq:HDR_LTIP_Complet}
    \begin{array}{c}
        \mathbf{f}_{HDR}(k,l) = g\left( \frac{1}{\eta}\sum_{i=1}^N w\left(\mathbf{f}_i(k,l) \right)\cdot
                            \mathbf{E}_i(k,l) \right) \\
                     = \frac{1}{\eta} \otimes
                        g\left(\sum_{i=1}^N w\left(\mathbf{f}_i(k,l) \right)\cdot  \mathbf{E}_i(k,l) \right) \\
                     = \frac{1}{\eta}  \otimes \left( \oplus \sum_{i=1}^N  g\left(w\left(\mathbf{f}_i(k,l) \right)\cdot  \mathbf{E}_i(k,l) \right) \right)\\
                    = \frac{1}{\eta} \otimes
                       \left( \oplus \sum_{i=1}^N \left(w\left(\mathbf{f}(k,l)\right) \right)\otimes  g\left(\mathbf{E}_i(k,l) \right) \right)\\
                    = \frac{1}{\eta} \otimes
                        \left( \oplus \sum_{i=1}^N \left(w\left(\mathbf{f}_i(k,l)\right) \right)\otimes
                        \mathbf{f}_i(k,l) \right)
    \end{array}
\end{equation}
where $\otimes$ and $\oplus$ are the LTIP operations shown in Table \ref{Tab:LTIPS}, while $\left(
\oplus \sum_{i=1}^N u_i\right)$ stands for:
\[
    \left( \oplus \sum_{i=1}^N u_i\right) = u_1 \oplus u_2 \oplus \dots  \oplus u_N
\]

Eq. (\ref{Eq:HDR_LTIP_Complet}) shows that one may avoid the conversion of  the input images to
irradiance maps, as the HDR image may be simply computed using additions and scalar multiplications
in the logarithmic domain. Furthermore, we accentuate that Eq. (\ref{Eq:HDR_LTIP_Complet}), if
written with real-based operations, matches  the exposure fusion introduced in \cite{Mertens:2007};
yet we have started our calculus based on the irradiance maps fusion. Thus, the use of LTIP
operations creates a unifying framework for both approaches. In parallel, it adds partial
motivation, by compatibility with HVS, for the exposure fusion variant. The motivation is only
partial as the LTIP model follows only the global HVS transfer function and not the local
adaptation.

The weights, $w\left(\mathbf{f}(k,l) \right)$, should complement the global tone mapping by
performing local adaptation. In \cite{Mann:2001} these weights are determined by derivation of the
CRF, while in \cite{Mertens:2007} they are extracted as to properly encode contrast, saturation and
well-exposedness.  More precisely:
\begin{itemize}
    \item \emph{contrast} $w_C$ is determined by considering the response of Laplacian operators; this is a
    measure of the local contrast which exists in human perception as center-surround ganglion field organization.
    \item \emph{saturation} $w_S$ is computed as the standard deviation of the R, G and B values, at each
    pixel location. This component favors photographic effects since the normal consumers are more attracted to
    vivid images and has no straight-forward correspondence to human perception.
    \item \emph{well-exposedness} $w_e$ is computed by giving small weights to values in
    the mid-range and large weights to outliers favoring the glistening aspect of consumer
    approaches. More precisely, one assumes that a perfect image is modelled by a Gaussian
    histogram with $\mu$ mean and $\sigma^2$ variance, and the weight of each pixel is the back--projected
    probability of its intensity given the named Gaussian.
\end{itemize}
We will assume the same  procedure of computing the weights with some small adjustments: while in
\cite{Mertens:2007} for well-exposedness both outliers were weighted symmetrically, we favor darker
tones to compensate the tendency of LIP models to favor bright tones, caused by their closing
property. Details about the precise implementation parameters values will be provided in Section
\ref{Sect:Results}, subsect. 1.


\section{Implementation and evaluation procedure}
\label{Sect:Implement}

\paragraph{Implementation} We have implemented the HDR algorithm described mainly by the
equation (\ref{Eq:HDR_LTIP_Complet}) within the LTIP model and weights similar to the procedure
described in \cite{Mertens:2007} in Matlab. The actual values of the weights are standard for
contrast $w_C=1$, saturation $w_S=1$, but differs for well-exposedness, where the mid range (the
parameter of the Gaussian distribution modelling it) are $\mu=0.37$ and $\sigma^2=0.2$. The choices
are based on maximizing the objective metrics and will be further explained in Sections
\ref{Sect:Results}.1 and \ref{Sect:Results}.3.

\begin{figure*}[t]
\begin{center}
    \begin{tabular}{ccc ccc}
        \includegraphics[width=0.14\linewidth]{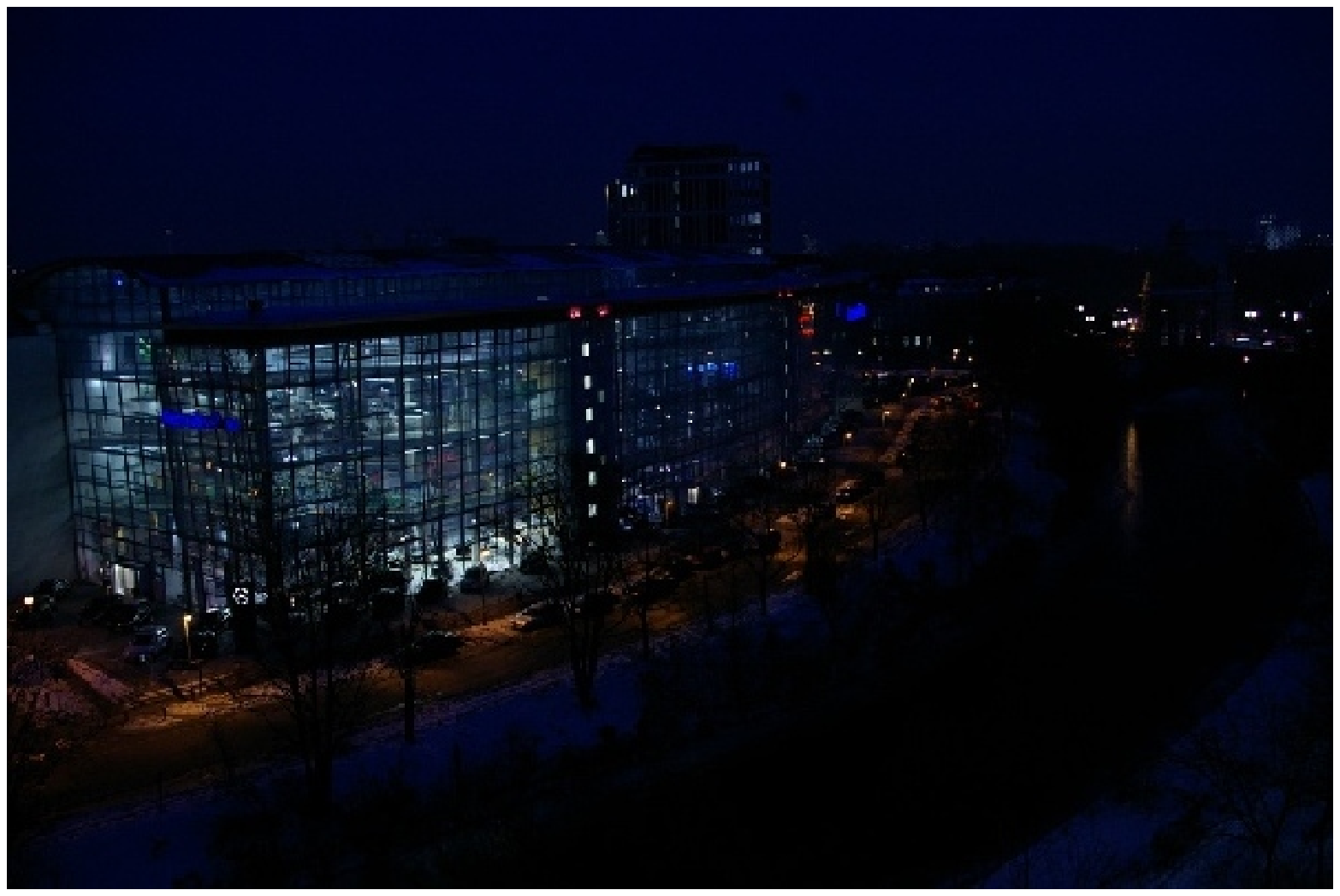} &
        \includegraphics[width=0.14\linewidth]{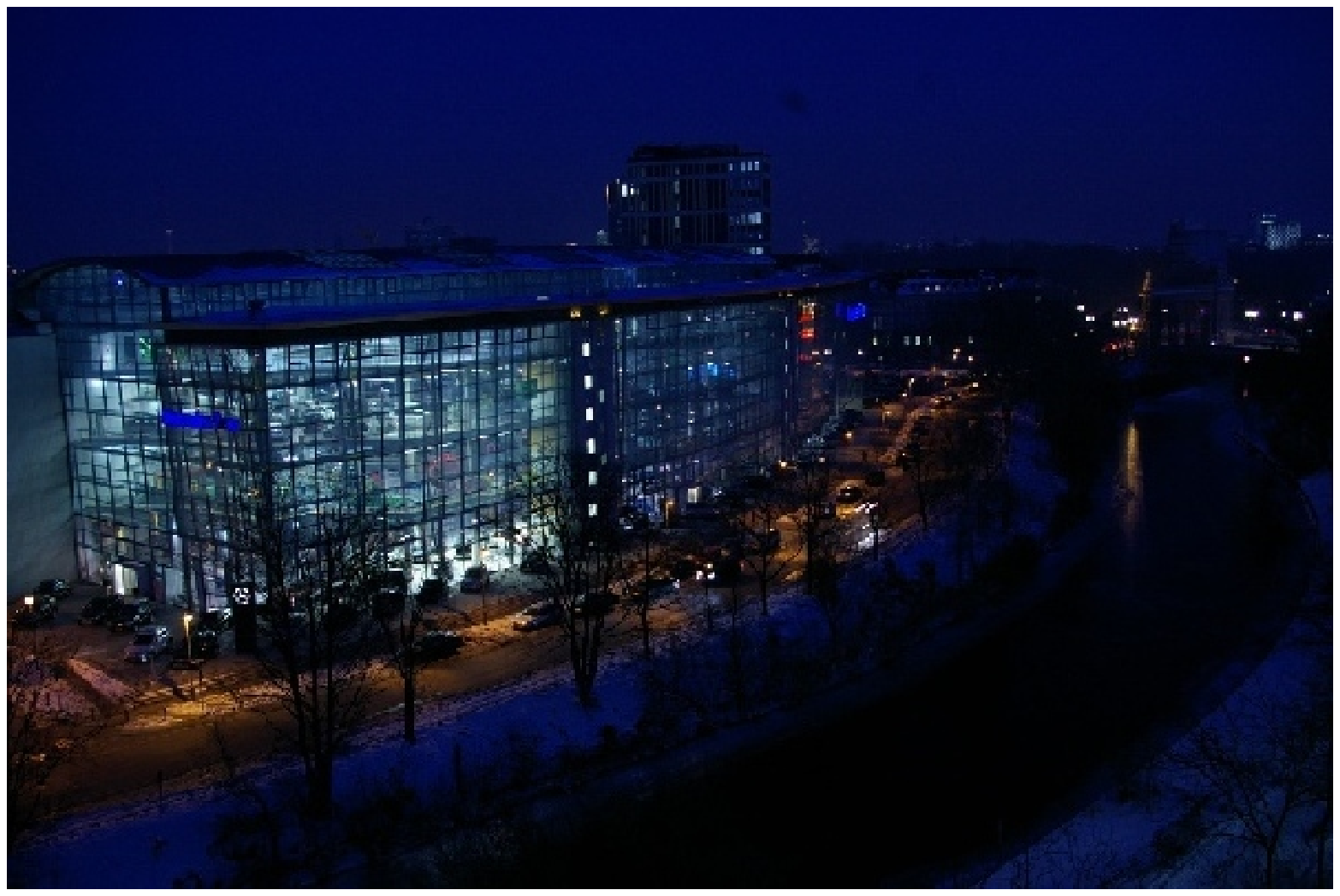} &
        \includegraphics[width=0.14\linewidth]{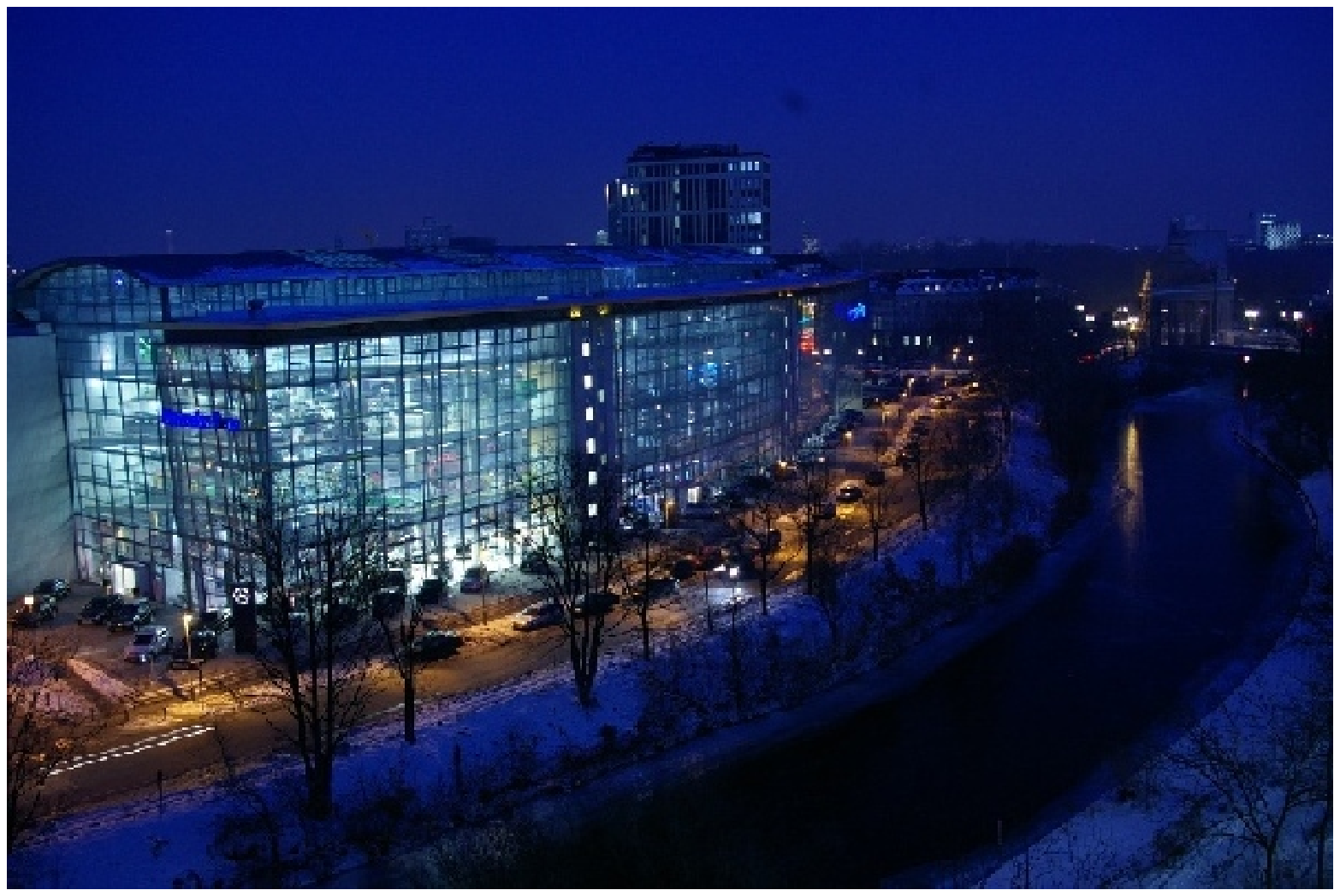} &
        \includegraphics[width=0.14\linewidth]{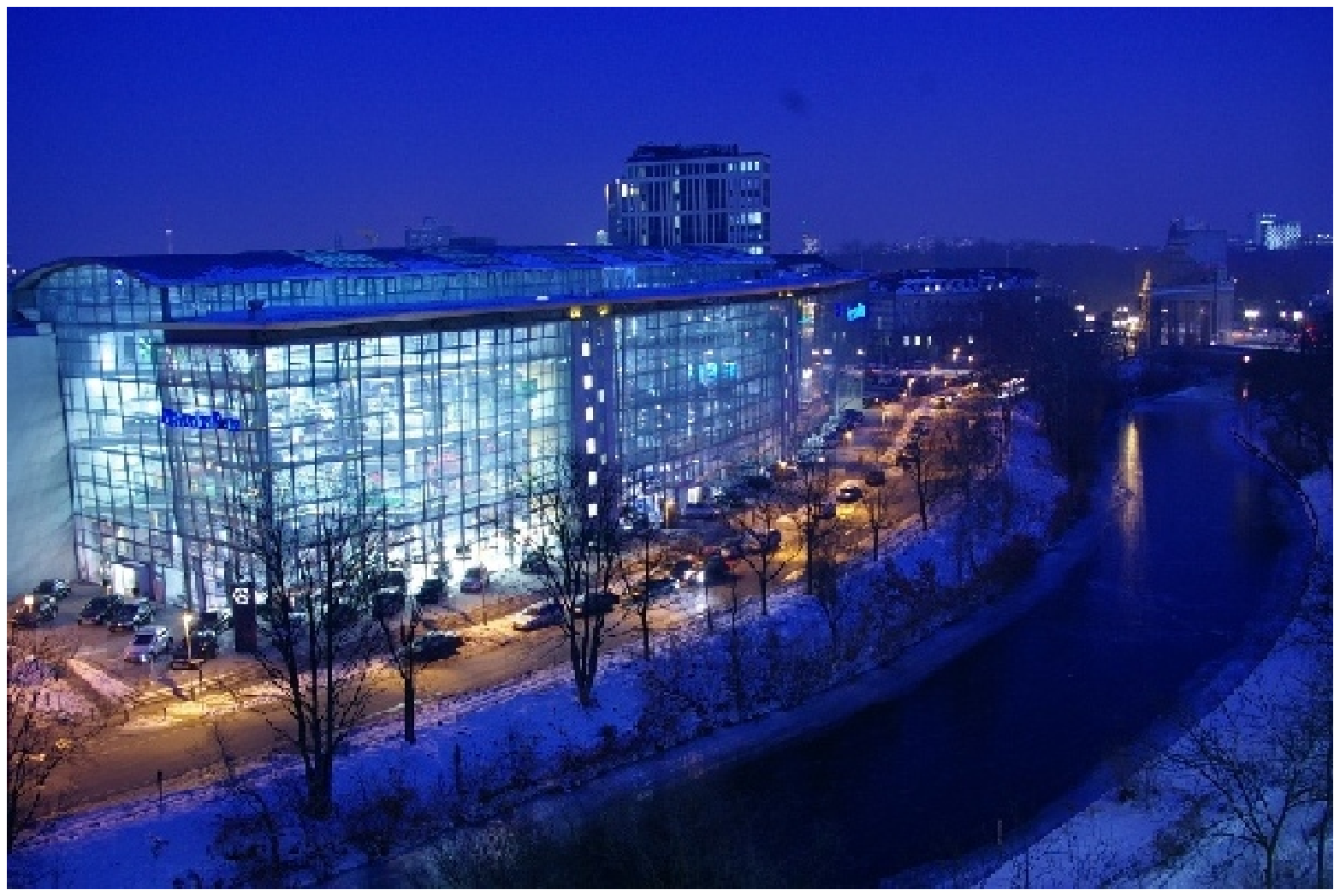} &
        \includegraphics[width=0.14\linewidth]{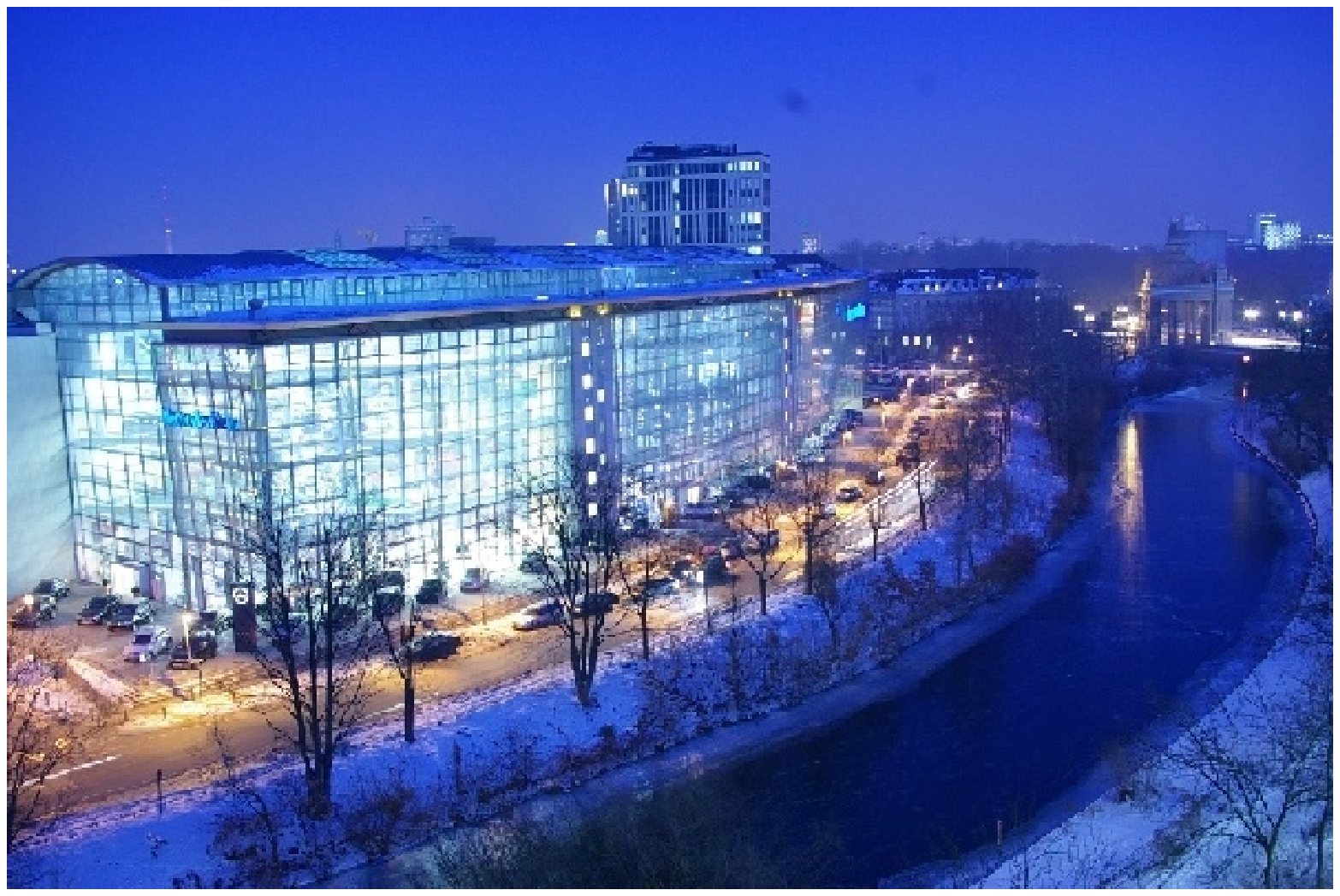} &
        \includegraphics[width=0.14\linewidth]{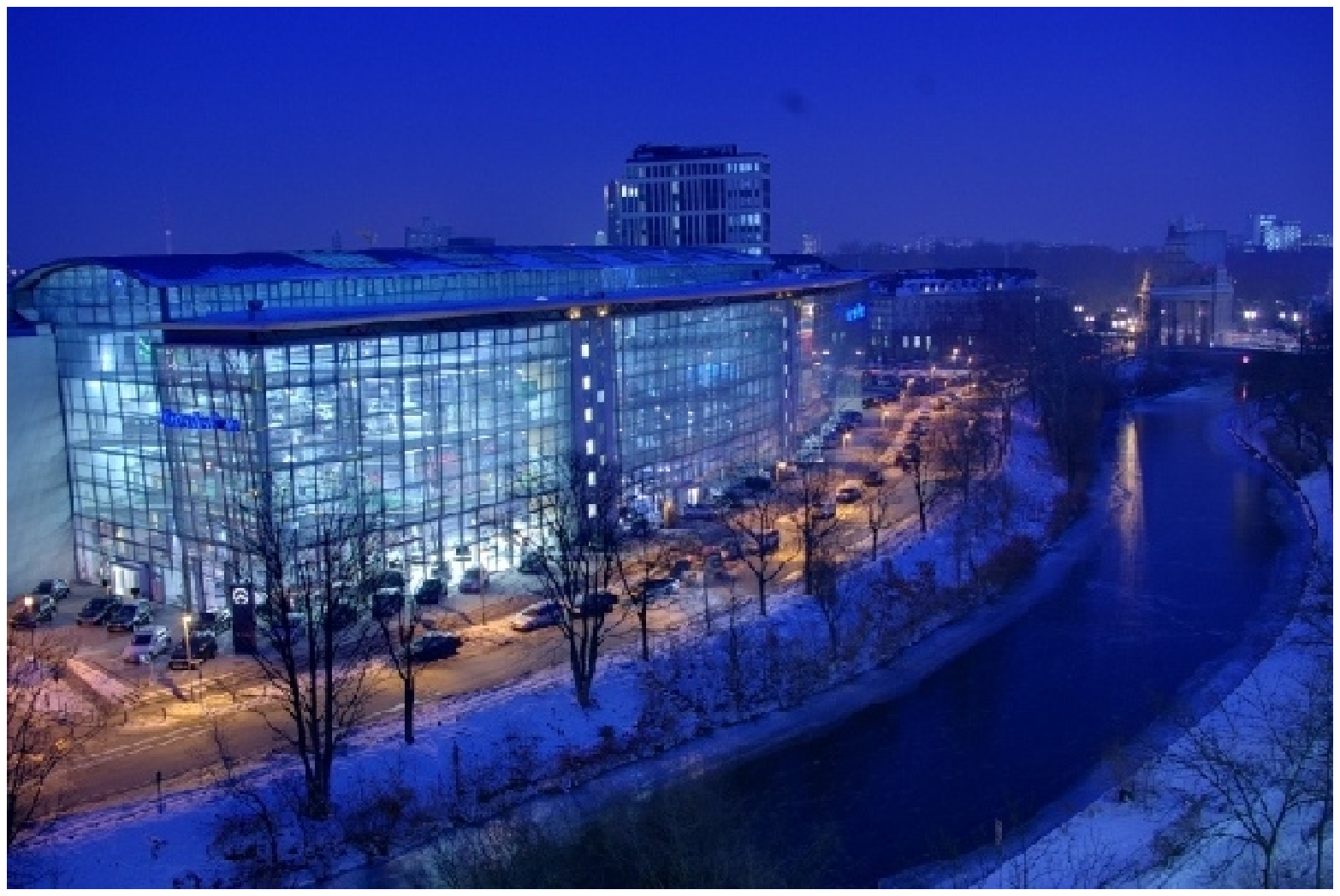}        \\
                (a) & (b) & (c) & (d) & (e) & (f) HDR
    \end{tabular}
    \caption{Example of HDR imaging: initial, differently exposed frames (a-e)) and the HDR image
    obtained by using the proposed algorithm (f).}
    \label{Fig:HDR_result}
\end{center}
\end{figure*}

An example of the achieved extended dynamic range image may be seen in Fig. \ref{Fig:HDR_result}.

The common practice is to evaluate HDR methods using few publicly available images. We adopted the
same principle, using more extensive public imagery data, such as the ones from \cite{Cadik:2008},
OpenCV examples library and from \cite{Drago:2003}. We have evaluated the proposed algorithm on a
database containing 22 sets of HDR frames acquired from various Internet sources, being constrained
by the fact that the proposed method requires also the original frames and not only the HDR image.
We made public\footnote{The code and supplementary results are available at
\url{http://imag.pub.ro/common/staff/cflorea/LIP} } the full results and the code to obtain them so
to encourage other people to further test it.

\paragraph{Evaluation} The problem of evaluating HDR images is still open as HDR techniques includes two categories:
irradiance map fusion which aims at correctness and exposure fusion which aims at pleasantness. As
mentioned in Section \ref{Sect:HDR_state}.2, the irradiance map is physically supported and typical
evaluation is performed with objective metrics that are inspired from human perception. Thus the
evaluation with such objective metrics will show how realistic is one method (i.e. how closely is
the produced image to the human perception of the scene).

On the other hand, the exposure fusion methods inspired from \cite{Mertens:2007} are much simpler,
and produce results without physically motivation, but which are visually pleasant for the average
user; consumer applications further process these image to enhance the surreal effects which is
deemed, but fake. Thus, the subjective evaluation and no-reference objective metrics that evaluate
the overall appearance will positively appreciate such images, although they are not a realistic
reproduction of the scene.

Thus, to have a complete understanding of a method performance, we will evaluate the achieved
results with two categories of methods: subjective evaluation and evaluation based on objective
metrics.


\subsection{Objective Evaluation}
While not unanimously accepted, several metrics were created for the evaluation of TMOs in
particular and HDR images in general. Here, we will refer to the metrics introduced in
\cite{Aydin:2008} and respectively the more recent one from \cite{Yeganeh:2013}.

The metric from \cite{Aydin:2008}, entitled ``Dynamic Range (In)dependent Image Metrics'' (DRIM)
uses a specific model of the HVS to construct a virtual low dynamic range (LDR) image from the HDR
reference and compares the contrast of the subject LDR image to the virtual one. In fact, the HVS
model and the comparison can be merged together, so that the matching is between the subject LDR
image and the HDR reference, skipping the virtual LDR image. The comparison takes into
consideration three categories: artificial amplification of contrast, artificial loss of contrast
and reversal of contrast. The metric points to pixels that are different from their standard
perception according to the authors aforethought HVS modelling and a typical monitor setting
($\gamma = 2.2$, 30 pixels per degree and viewing distance of 0.5m). For each test image we
normalized the error image by the original image size (as described in \cite{Ferradans:2011}). The
metric only assigns one type of error (the predominant one) and has two shortcomings: it heavily
penalizes global amplification error (which is not so disturbing from a subjective point of view)
and it merely penalizes artifacts (such as areas with completely wrong luminance), which, for a
normal viewer, are extremely annoying. Thus the metric, in fact, assigns a degree of
\emph{perceptualness} (in the sense of how close is that method to the human contrast transfer
function) to a certain HDR method.

A more robust procedure for evaluation was  proposed in \cite{Yeganeh:2013} where, in fact, three
scores are introduced:
\begin{itemize}
    \item \emph{Structural fidelity}, $S$, which  uses the structural similarity image metric (SSIM)
    \cite{Wang:04} to establish differences between the LDR image and the original HDR one and a bank of
    non-linear filters based on the human contrast sensitivity function (CSF) \cite{Barten:1999}. The
    metric points to structural artifacts of the LDR image with respect to the HDR image  and has a 0.7912 correlation with
    subjective evaluation according to Yeganeh et al. \yrcite{Yeganeh:2013}.
    \item \emph{Statistical naturalness}, $N$ gives a score of the closeness of the image histogram to a normal
    distribution which was found to match average human opinion.
    \item \emph{Overall quality}, $Q$ which integrates the structural fidelity, $S$, and statistical
    naturalness, $N$, by :
    \begin{equation}
        Q = aS^\alpha + (1 - a)N^\beta
    \end{equation}
     where $a = 0.8012$, $\alpha = 0.3046$,  $\beta = 0.7088$ as found in \cite{Yeganeh:2013}. The
     metric has 0.818 correlation with human opinion.
\end{itemize}

The structural fidelity  appreciates how close a TMO to the CSF function, thus is theoretical
oriented measure, while the structural fidelity is subjective measure and shows how close is a TMO
to the consumer preferences.


\subsection{Subjective evaluation}

On the subjective evaluation, for HDR images, {\v{C}}ad\'{i}k et al. \cite{Cadik:2008} indicated
the following criteria as being relevant: luminosity, contrast, color and detail reproduction, and
lack of image artifacts. The evaluation  was performed in two steps. First we analyzed
comparatively, by means of example, the overall appearance and existence of artifacts with respect
to the five named criteria in the tested methods;  next, we followed with subjective evaluation
where an external group of persons graded the images.

To perform the external evaluation, we instructed 18 students in the 20-24 years range to examine
and rank the images on their personal displays, taking into account the five named criteria. We
note that the students follow computer science or engineering programme, but they are not closely
related to image processing. Thus, the subjective evaluation could be biased towards groups with
technical expertise.

The testing was partially blind: the subjects were aware of the theme (i.e. HDR), but they were not
aware about the source of each image. While we introduced them to a method for monitor calibration
and discussed aspects about view angle and distance to monitor, we did not impose them as strict
requirements since the average consumer does not follow rigorous criteria for visualization. Thus,
the subjective evaluation was more related to how appealing an image is.


\section{Results}
\label{Sect:Results}
\subsection{Algorithm parameters}
To determine the best parameters of the proposed method we resorted to empirical validation.

\begin{figure*}[t]
    \begin{center}
    \begin{tabular}{cccc}
        \includegraphics[width=0.20\linewidth]{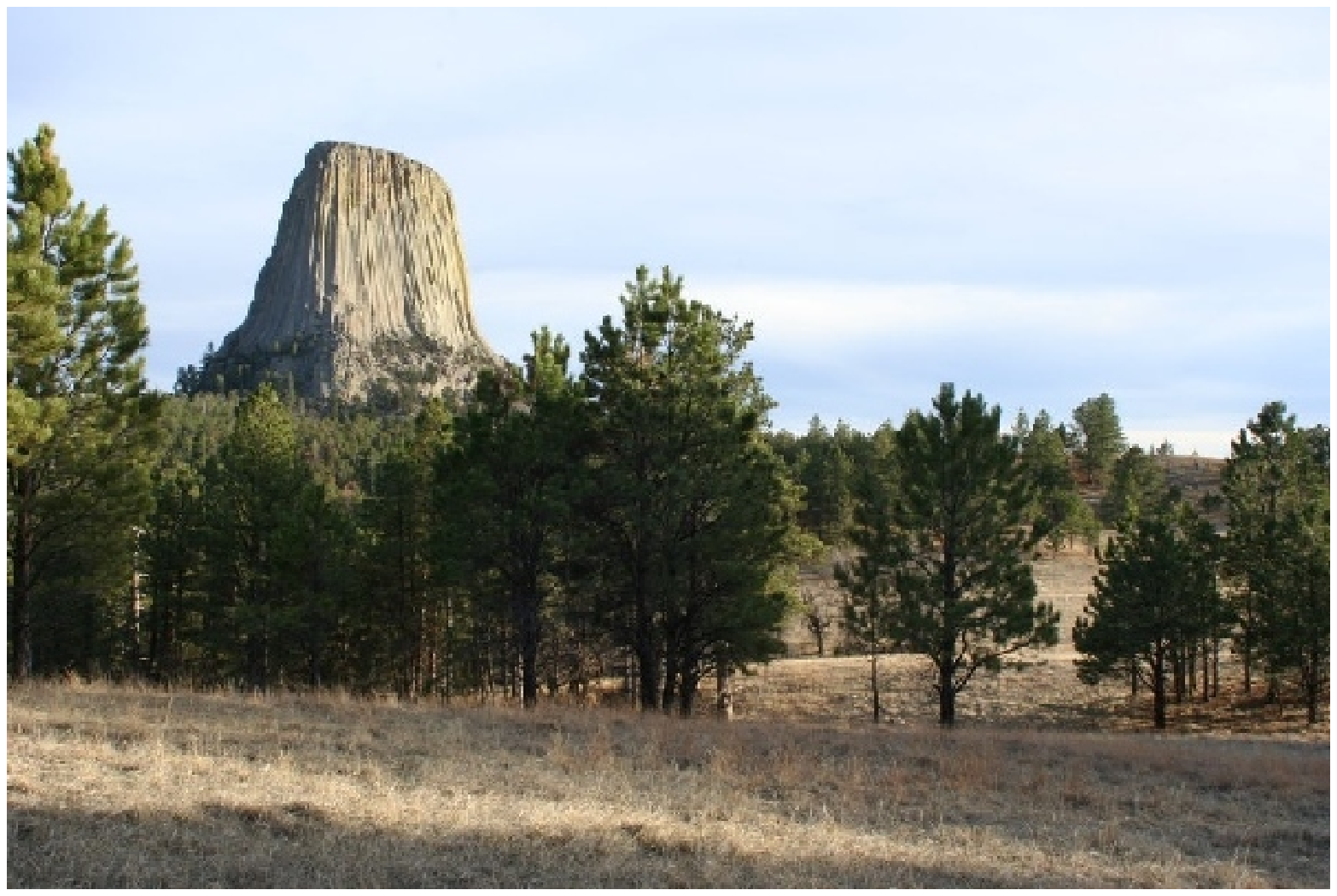} &
        \includegraphics[width=0.20\linewidth]{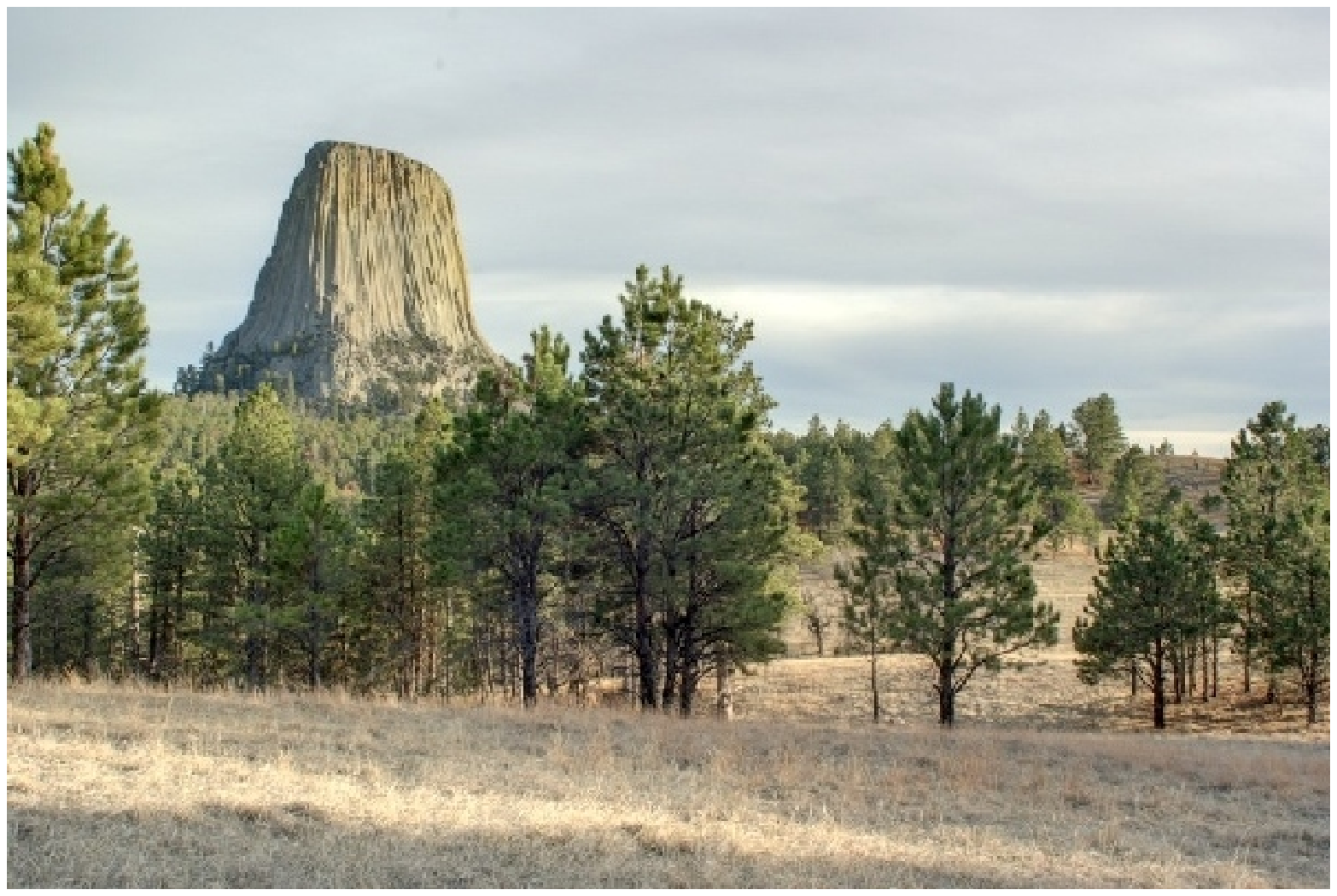} &
        \includegraphics[width=0.20\linewidth]{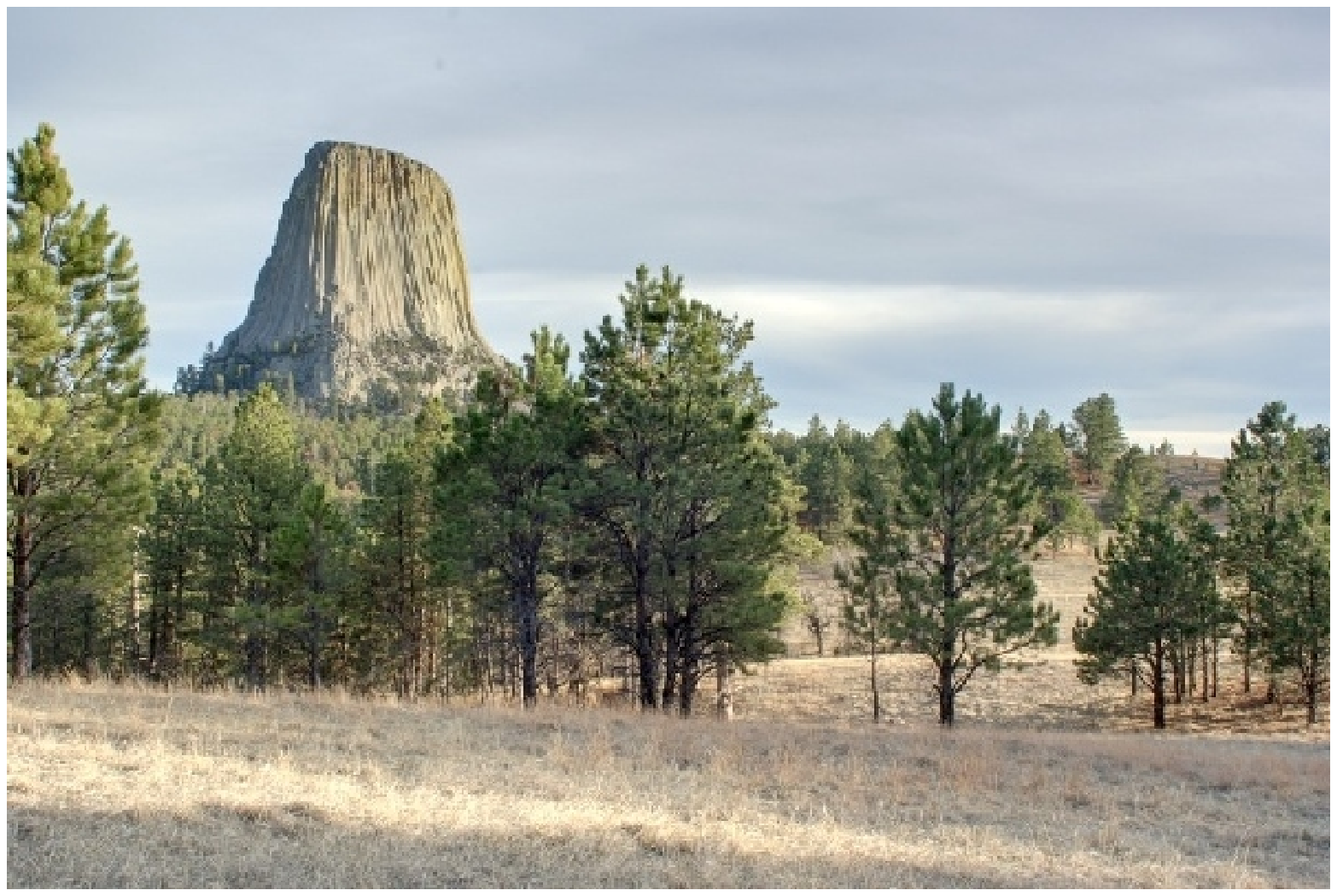} &
        \includegraphics[width=0.20\linewidth]{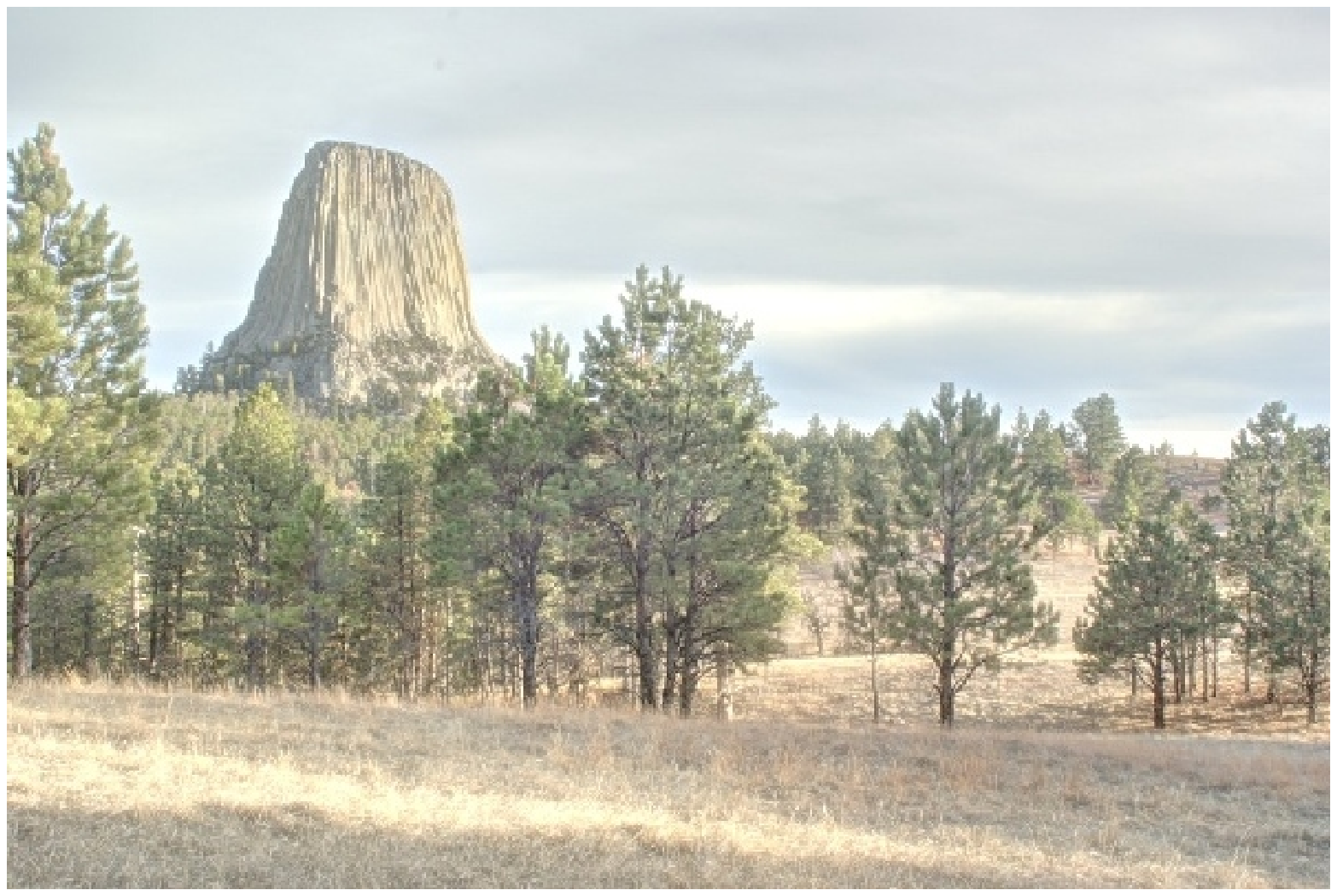} \\
        (a) Normally exposed frame & (b) LTIP, $m=1$ & (c) LIP & (d) symmetrical LIP \\
        \includegraphics[width=0.20\linewidth]{devilFusLTIP050.eps} &
        \includegraphics[width=0.20\linewidth]{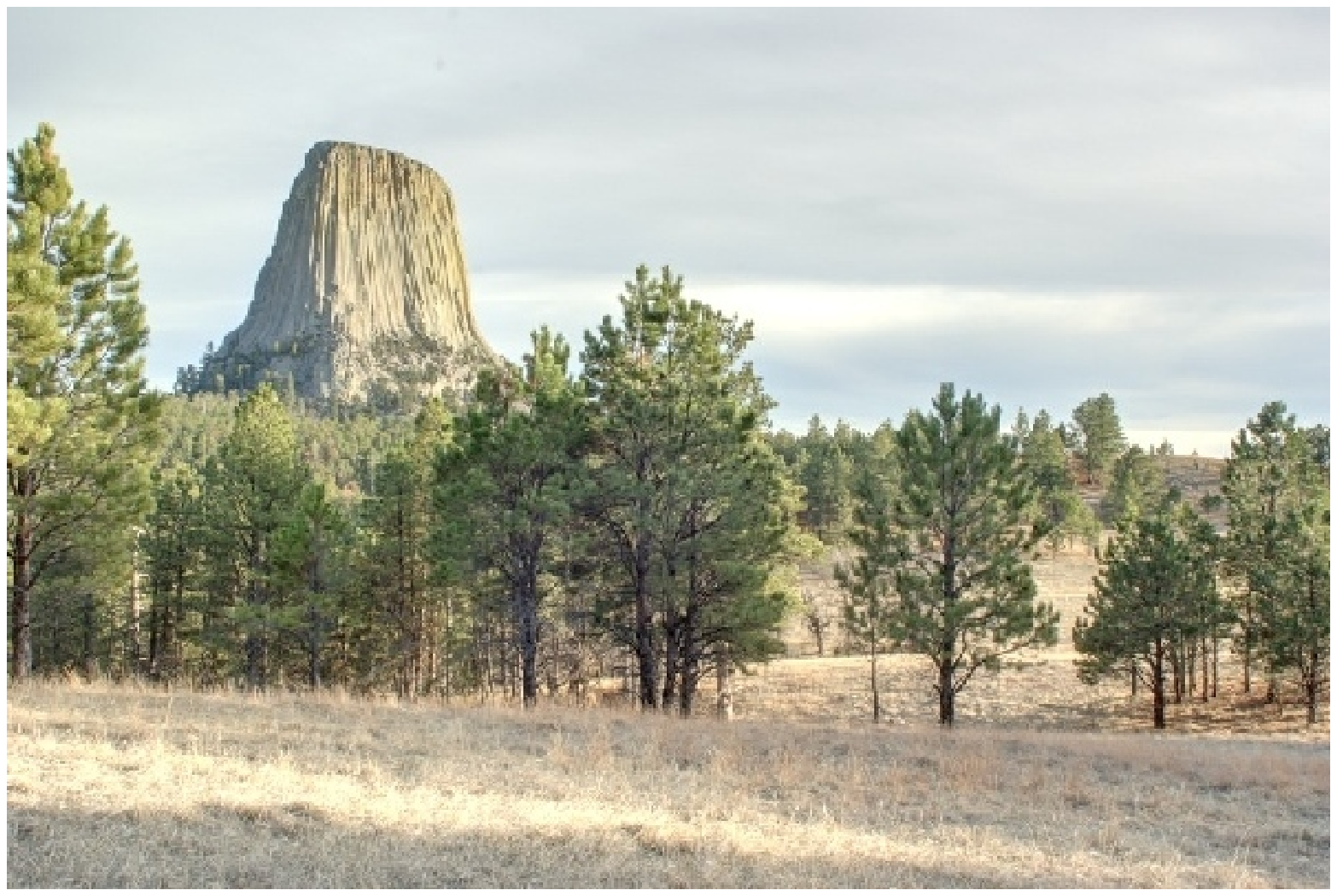} &
        \includegraphics[width=0.20\linewidth]{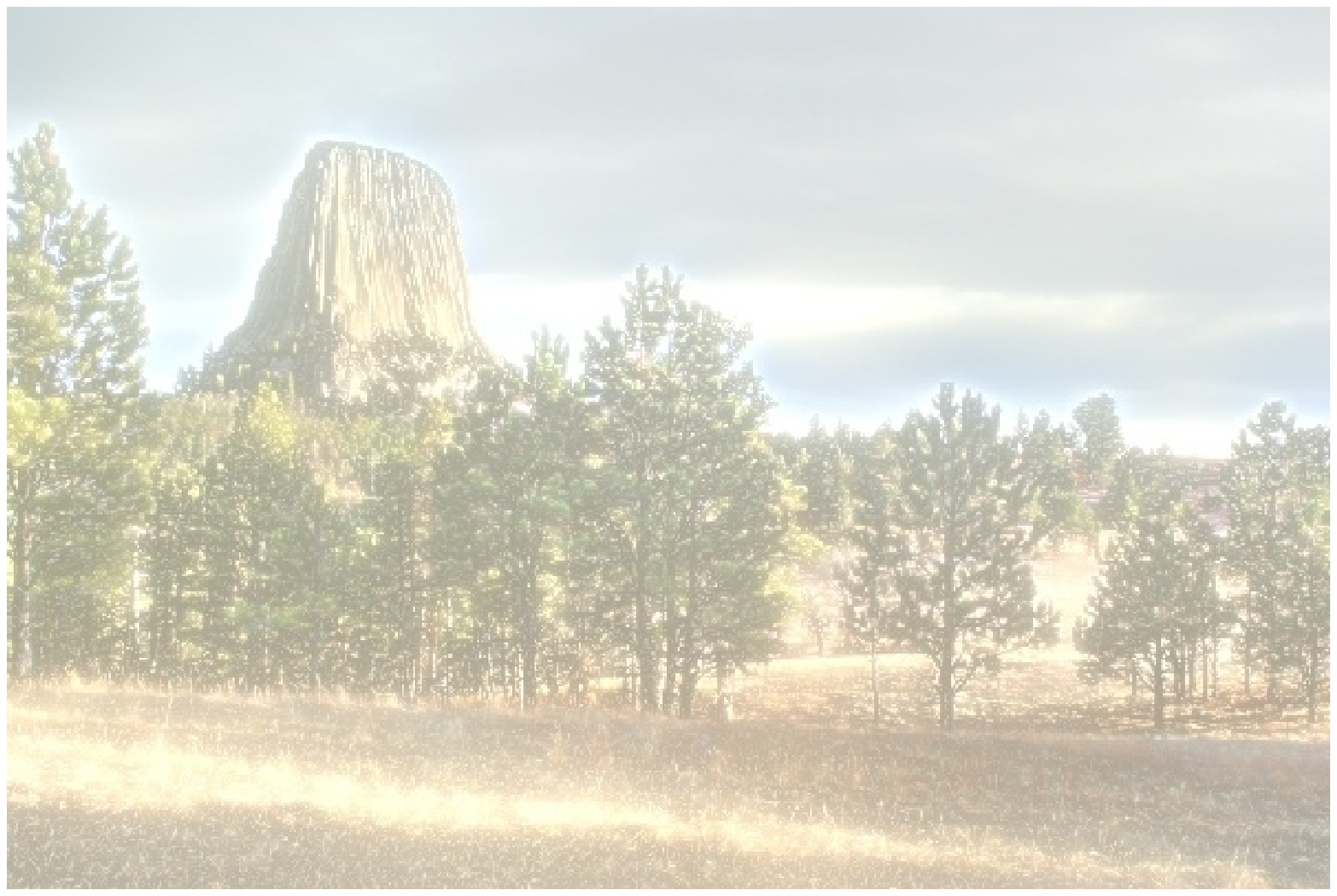} &
        \includegraphics[width=0.20\linewidth]{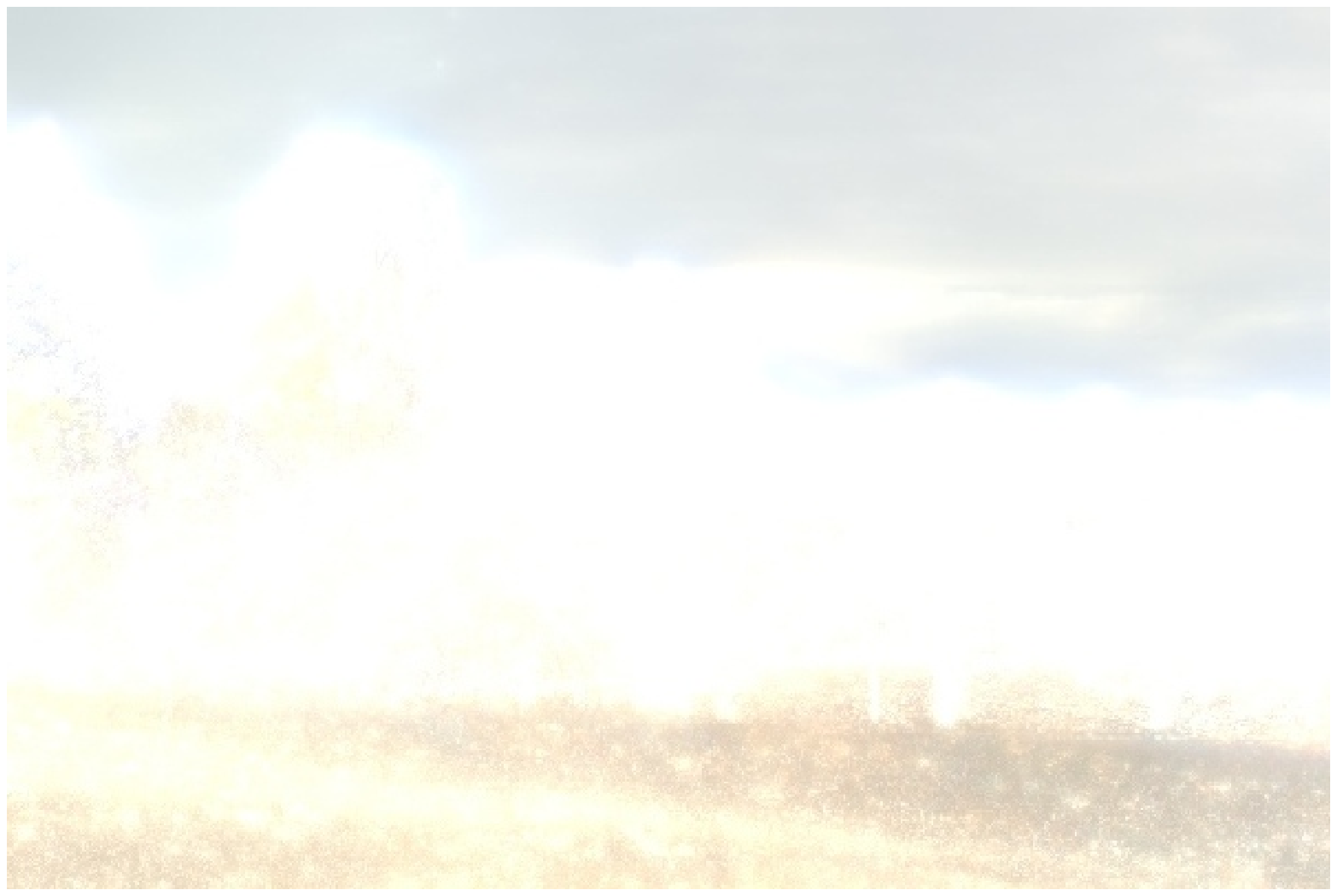}        \\
        (e) LTIP, $m = 0.5$  & (f) LTIP, $m = 0.75$ & (g) LTIP, $m = 1.33$ & (h) LTIP, $m = 2$
    \end{tabular}
    \caption{Normal exposed image frame (a) and the resulting image obtained by the standard LTIP
    model (b), Classical LIP model (c), symmetrical model  introduced by Patrascu (d). Images obtained
    with the parametric extension of the LTIP model (e-h). }
    \label{Fig:LIP_Examples}
    \end{center}
\end{figure*}

 \emph{Logarithmic Model}. The first choice of the proposed method is related
to the specific Logarithm Image Processing model used. While we discuss this aspect by means of an
example shown in Fig. \ref{Fig:LIP_Examples}, top row (b-d), we have to stress that all the results
fall in the same line. The LTIP model provides the best contrast, while the classical LIP model
\cite{Jour:1987} leads to very similar results, with marginal differences like slightly flatter sky
and less contrast on the forest. The symmetrical model \cite{Patrascu:2001} produces
over--saturated images. Given the choice between our solution and the one based on
\cite{Jour:1987}, as differences are rather small, the choice relies solely on the perceptual
motivation detailed in Section \ref{Sect:FLIPbasedTMO}.

Next, given the parametric extension of the LTIP model from \cite{Florea:2013}, we asked which is
the best value for the parameter $m$. As shown in Fig. \ref{Fig:LIP_Examples}, bottom row (e-h),
the best visual results are obtained for $m=1$, which corresponds to the original LTIP model.
Choices different from $m=1$ use direct or inverse transformations that are too concave and,
respectively, too convex, thus distorting the final results. Also, the formulas become increasingly
complex and precise computation more expensive. Concluding, the best results are achieved with
models that are closer to the human perception.

\emph{Algorithm weights}. In Section \ref{Sect:FLIPbasedTMO} we nominated three categories of
weights (contrast -- $w_C$, saturation $w_S$ and well-exposedness) that interfere with the
algorithm. For the first two categories, values different from standard ones ($w_C=1$ and $w_S=1$)
have little impact.

\begin{figure*}[t]
    \begin{center}
    \begin{tabular}{cccc}
        \includegraphics[width=0.20\linewidth]{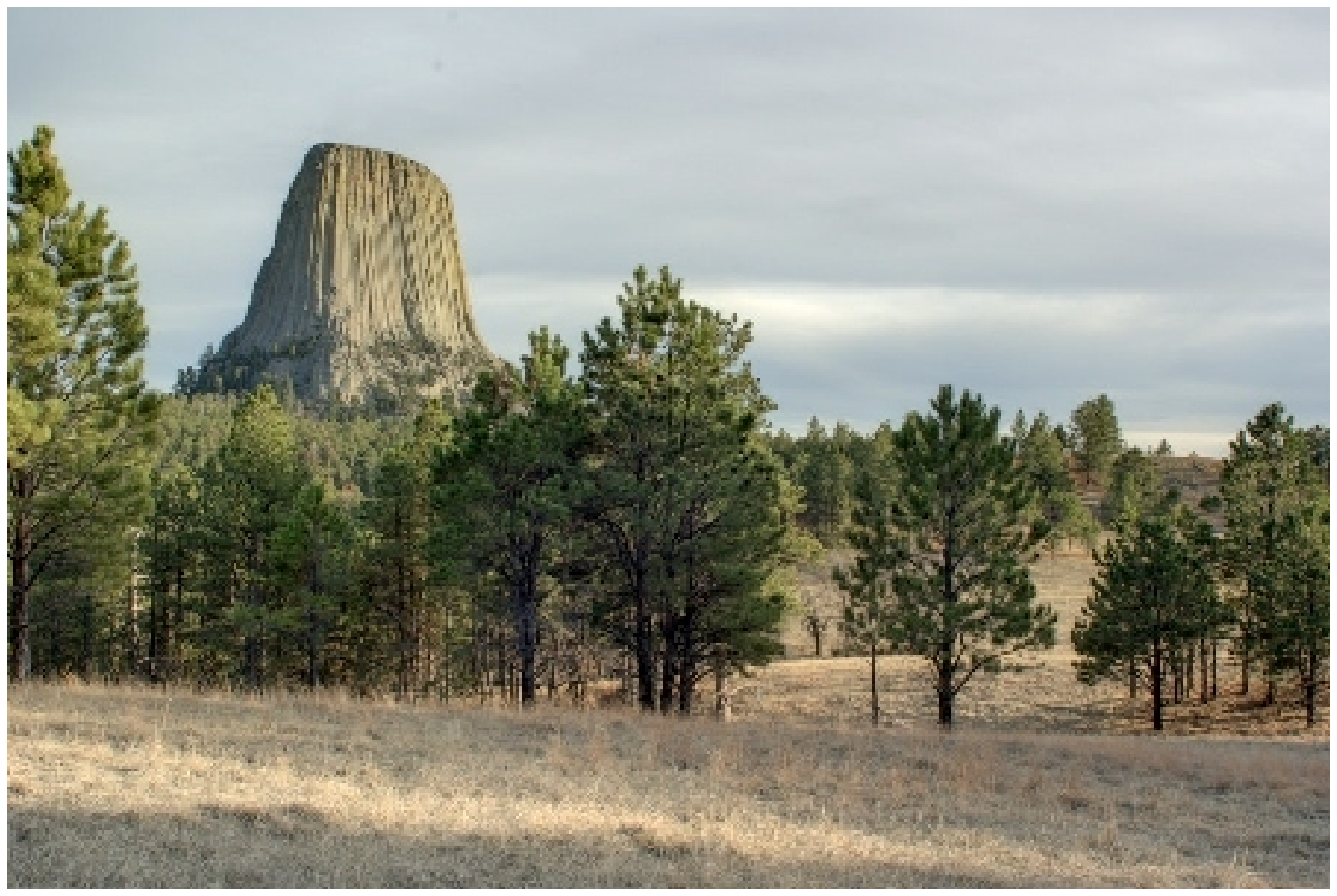} &
        \includegraphics[width=0.20\linewidth]{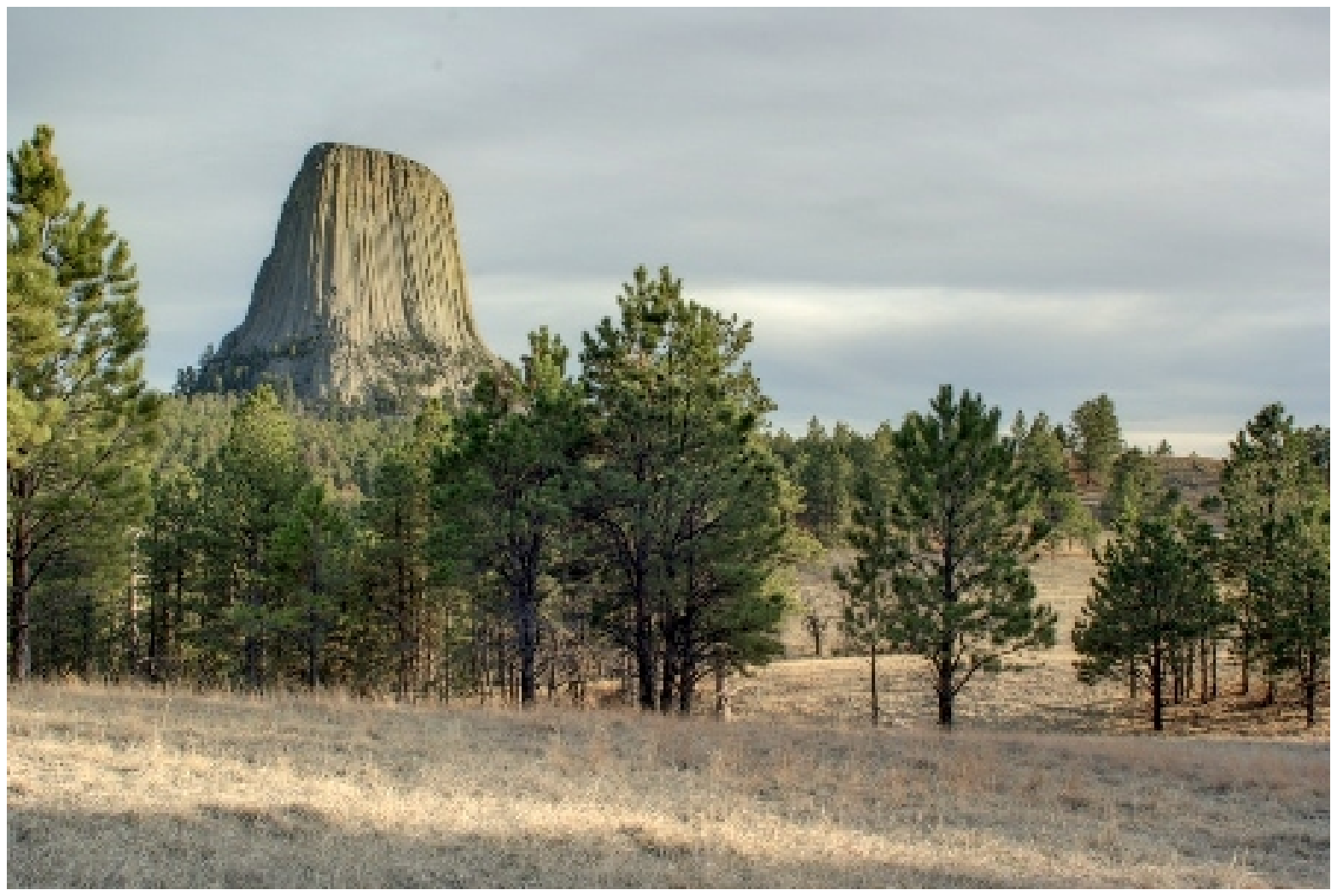} &
        \includegraphics[width=0.20\linewidth]{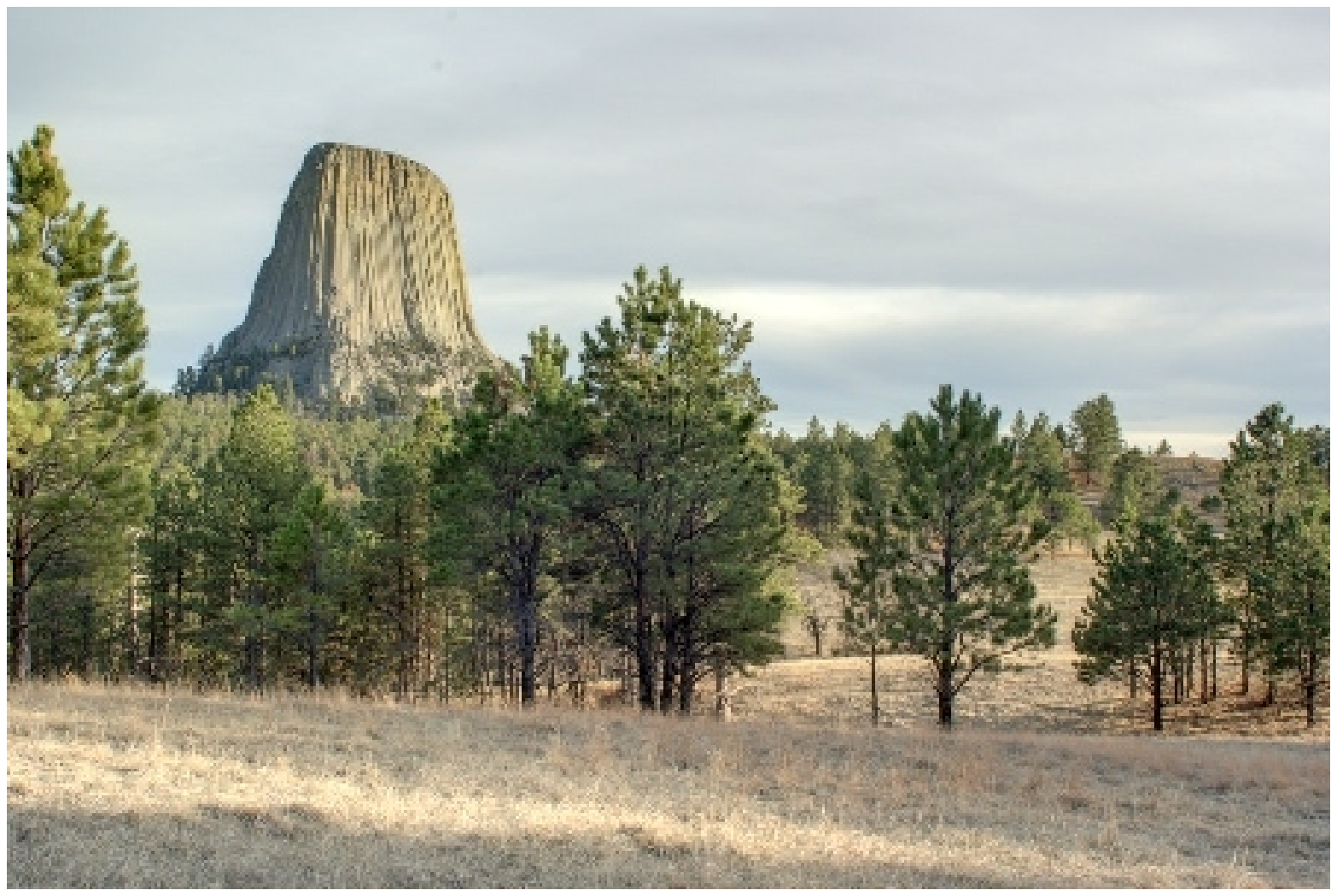} &
        \includegraphics[width=0.20\linewidth]{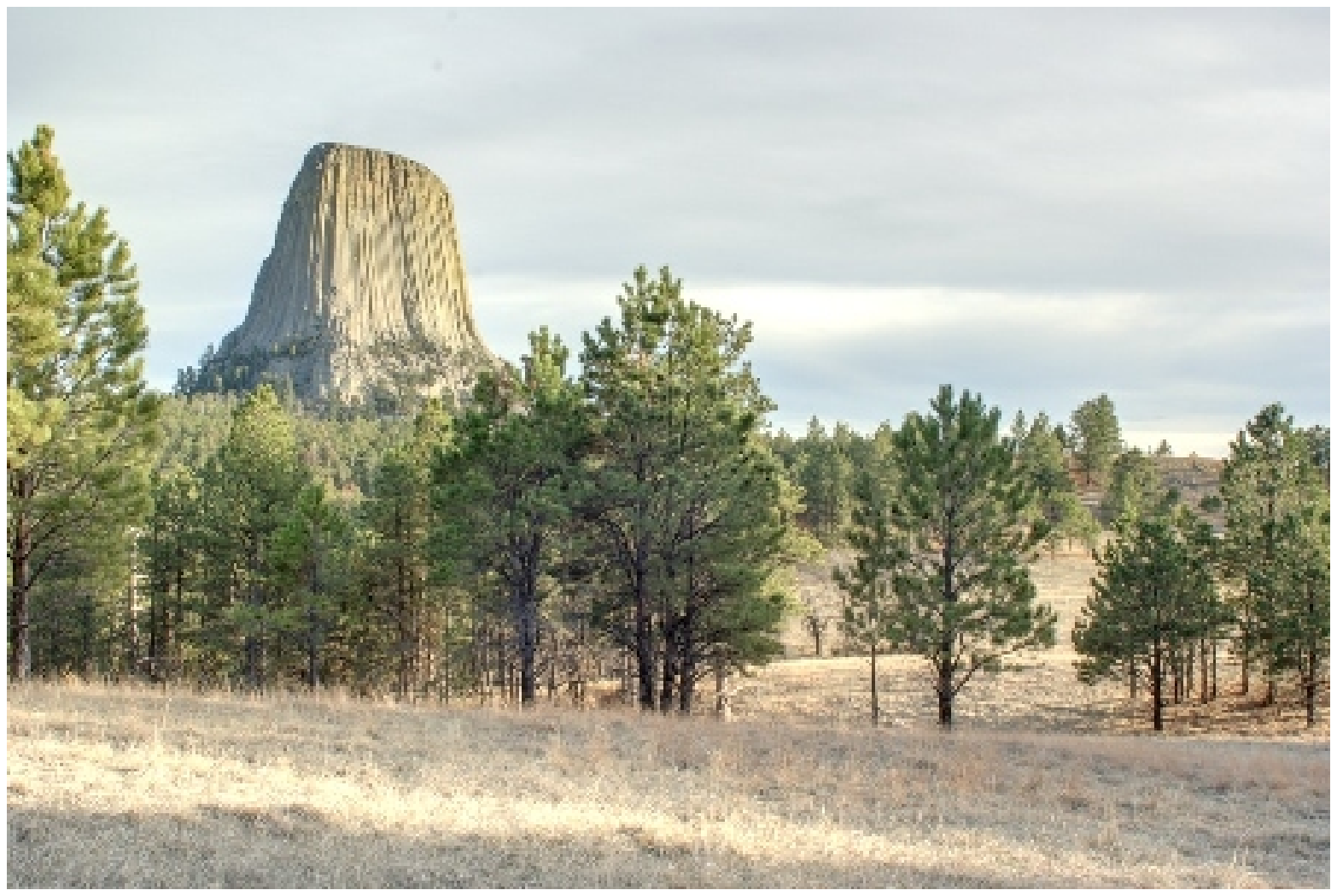} \\
        (a) $\mu=0.32$ & (b) $\mu=0.37$ & (c) $\mu=0.4$ & (d) $\mu=0.5$
    \end{tabular}
    \caption{Output image when various values for mid-range ($\mu$) in well-exposedness weight were
    used. The preferred choice is $\mu=0.37$. }
    \label{Fig:MuWeights}
    \end{center}
\end{figure*}

The well-exposedness, which is described by mainly the central value $\mu$ of the ''mid range'' has
significant impact. As one can see in Fig. \ref{Fig:MuWeights}, the best result is achieved for
$\mu=0.37$ while for larger values ($\mu>0.37$) the image is too bright and, respectively, for
smaller ones ($\mu<0.37$) is too dark. While a $\mu=0.4$ produced similar results, the objective
metrics reach the optimum in $\mu=037$. Changing the variance also has little impact. These
findings were confirmed by objective testing, as further showed in Section \ref{Sect:Results}.3.

\subsection{Comparison with state of the art}

To test against various state of the art methods, we used the HDR irradiance map (stored as
\texttt{.hdr} file) which was either delivered with the images (and typically produced using the
method from \cite{Robertson:1999}), or produced with some online available code\footnote{The HDR
creator package is available at \url{http://cybertron.cg.tu-berlin.de/pdci09/} $\phantom{XXXXX}$
\url{hdr_tonemapping/download.html}}.

\begin{figure*}[t]
    \begin{center}
    \begin{tabular}{cccc}
        \includegraphics[width=0.21\linewidth]{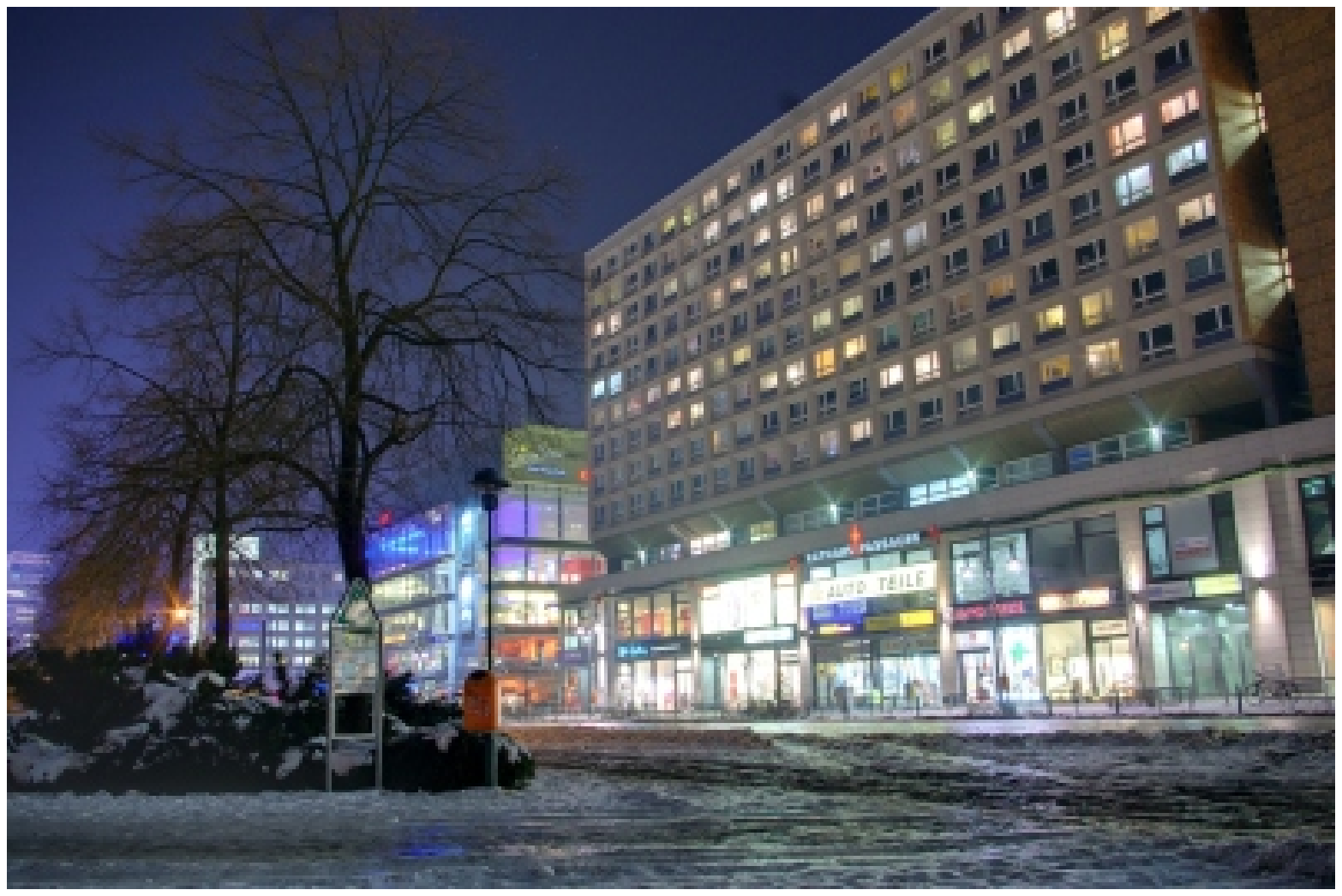} &
        \includegraphics[width=0.21\linewidth]{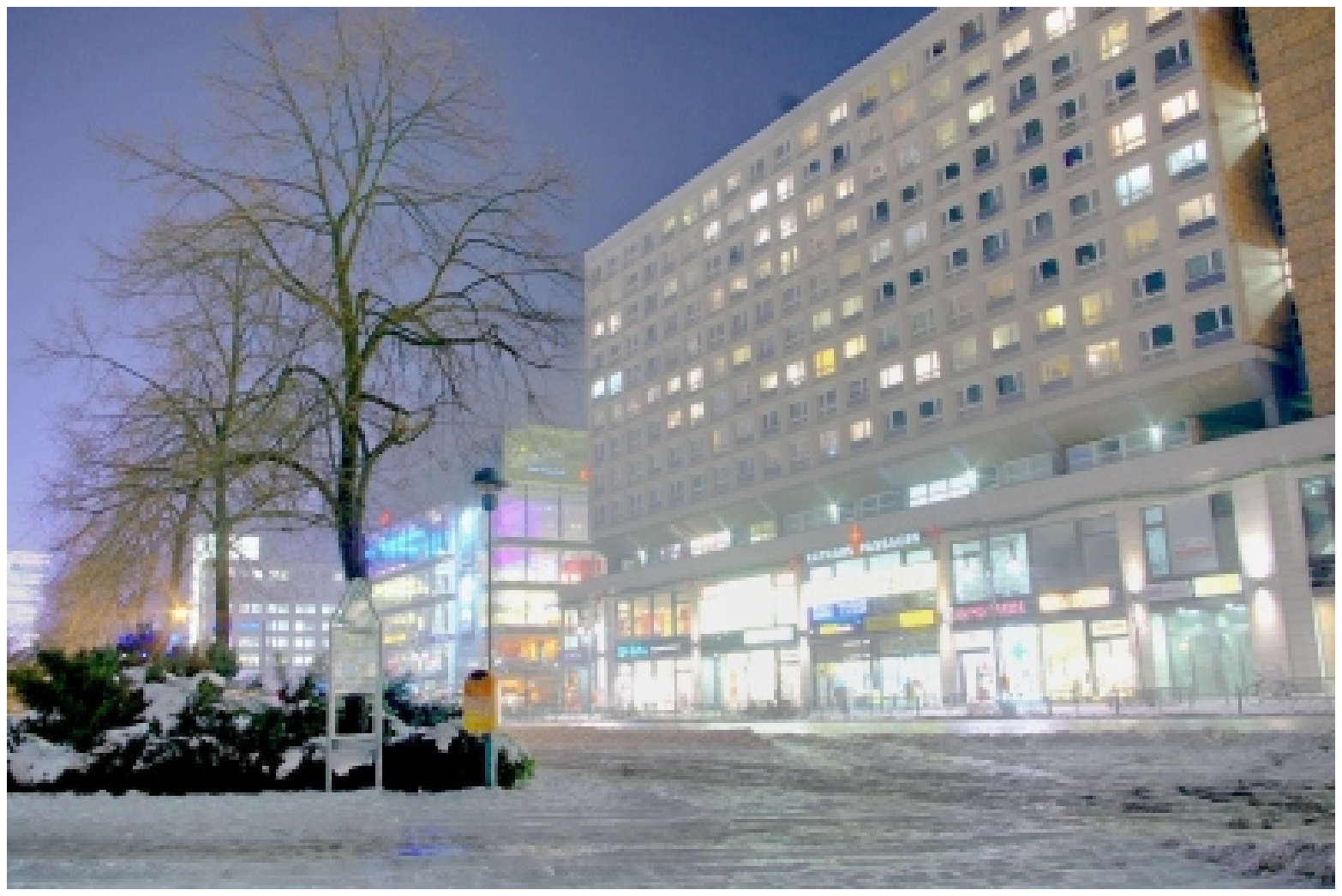} &
        \includegraphics[width=0.21\linewidth]{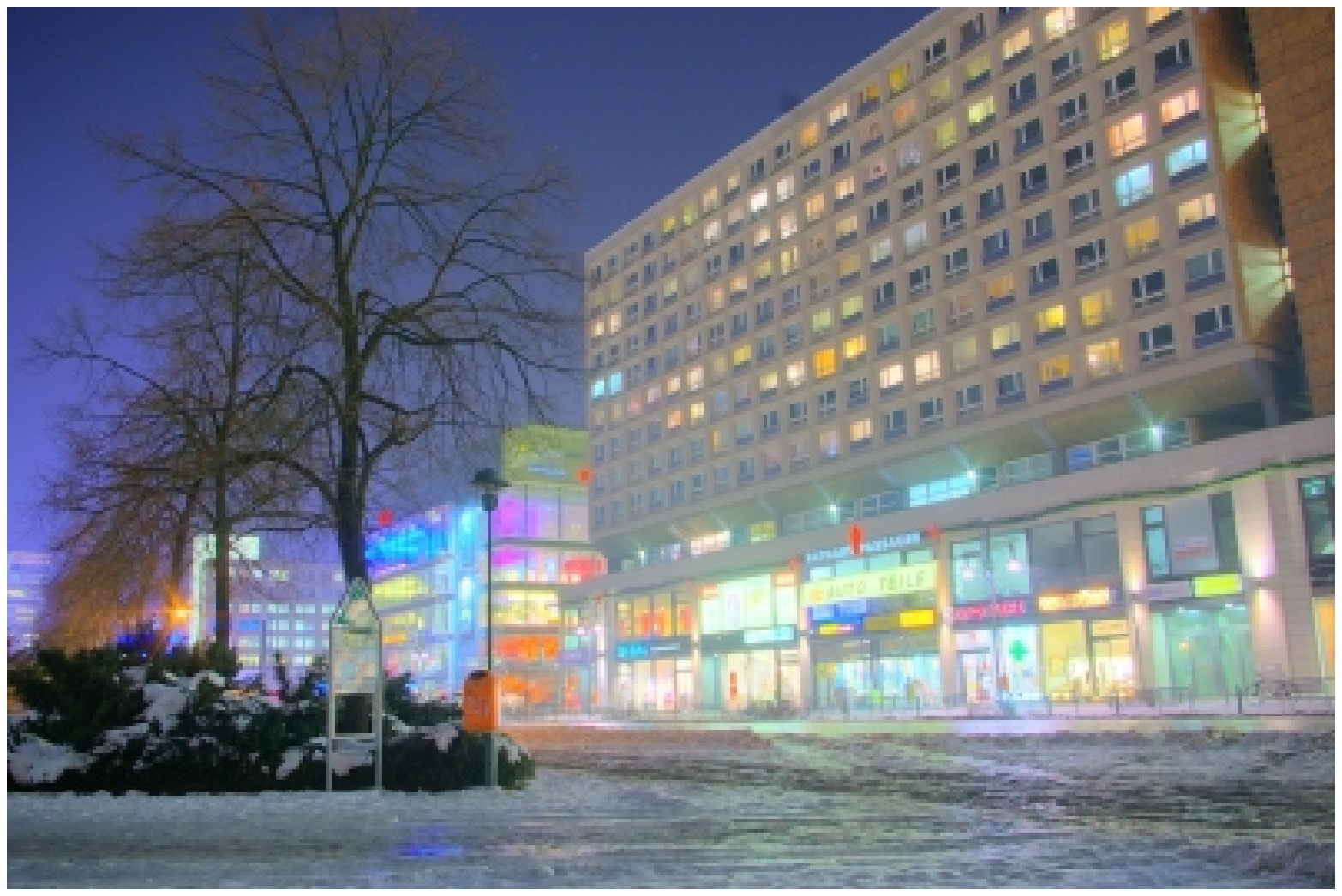} &
        \includegraphics[width=0.21\linewidth]{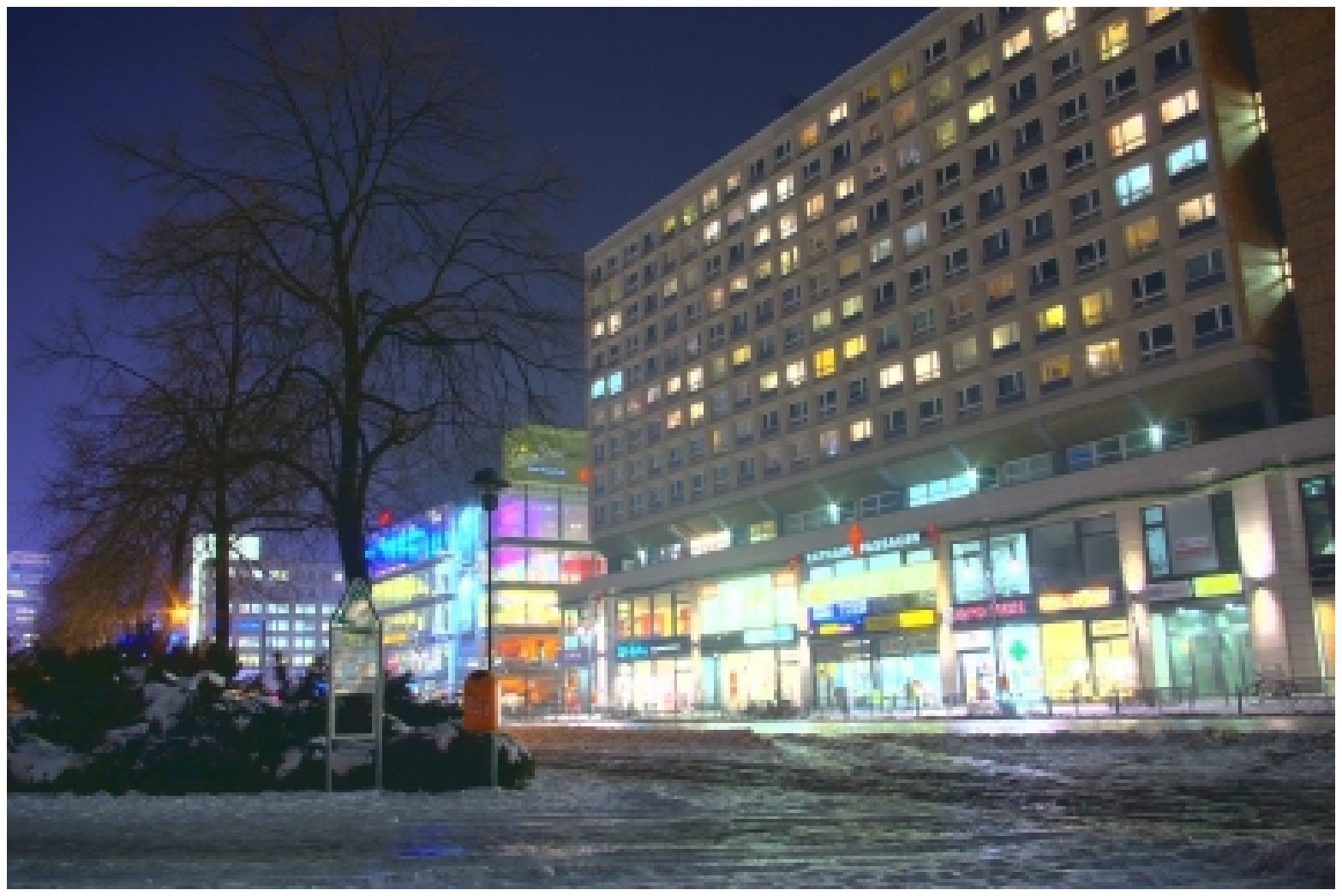} \\
            (a) \textbf{Proposed} & (b) Pece and Kautz \yrcite{Pece:2010} &
            (c) Banterle et al. \yrcite{Banterle:2012} & (d)Krawczyk et al. \yrcite{Krawczyk:2005} \\
        \includegraphics[width=0.21\linewidth]{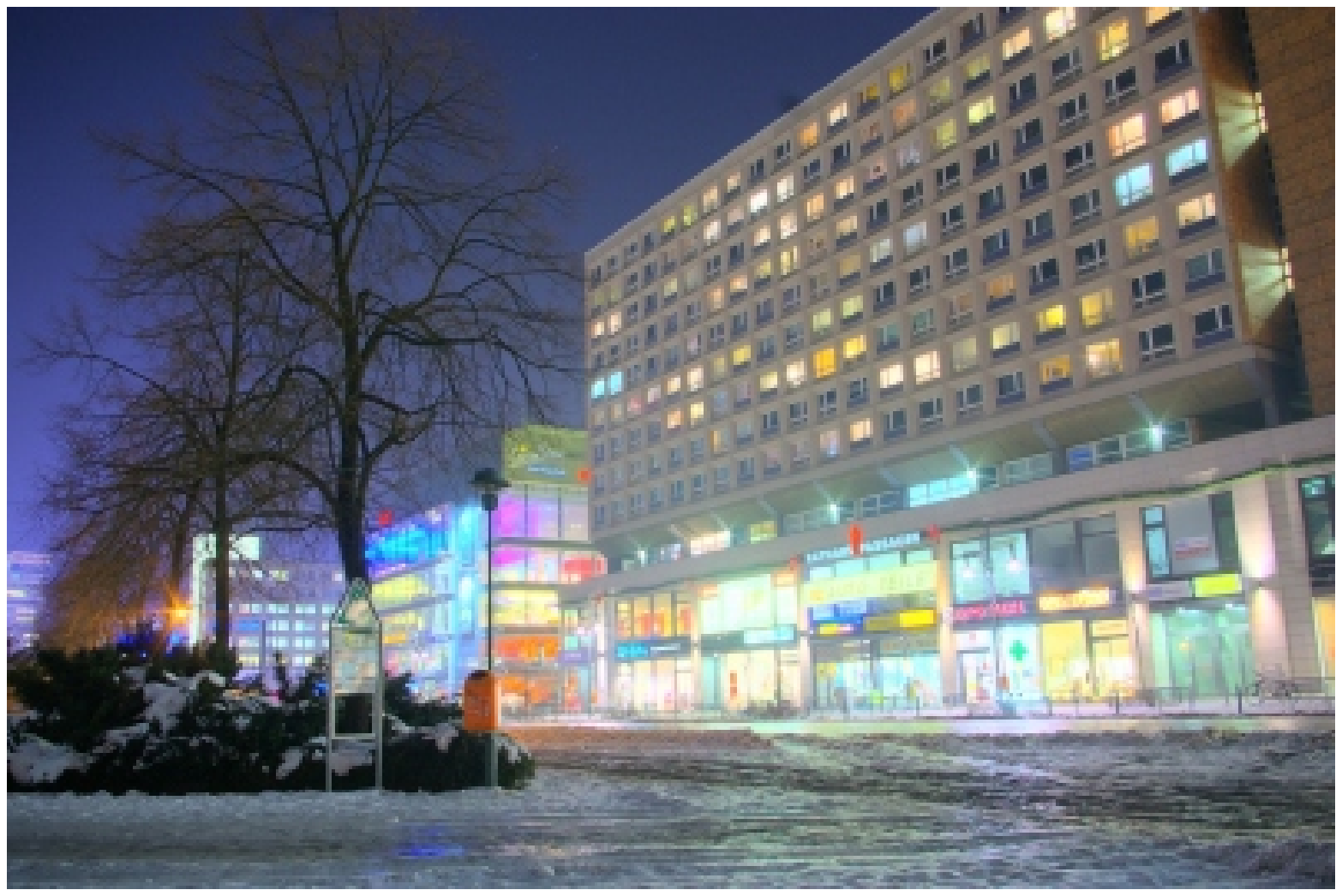} &
        \includegraphics[width=0.21\linewidth]{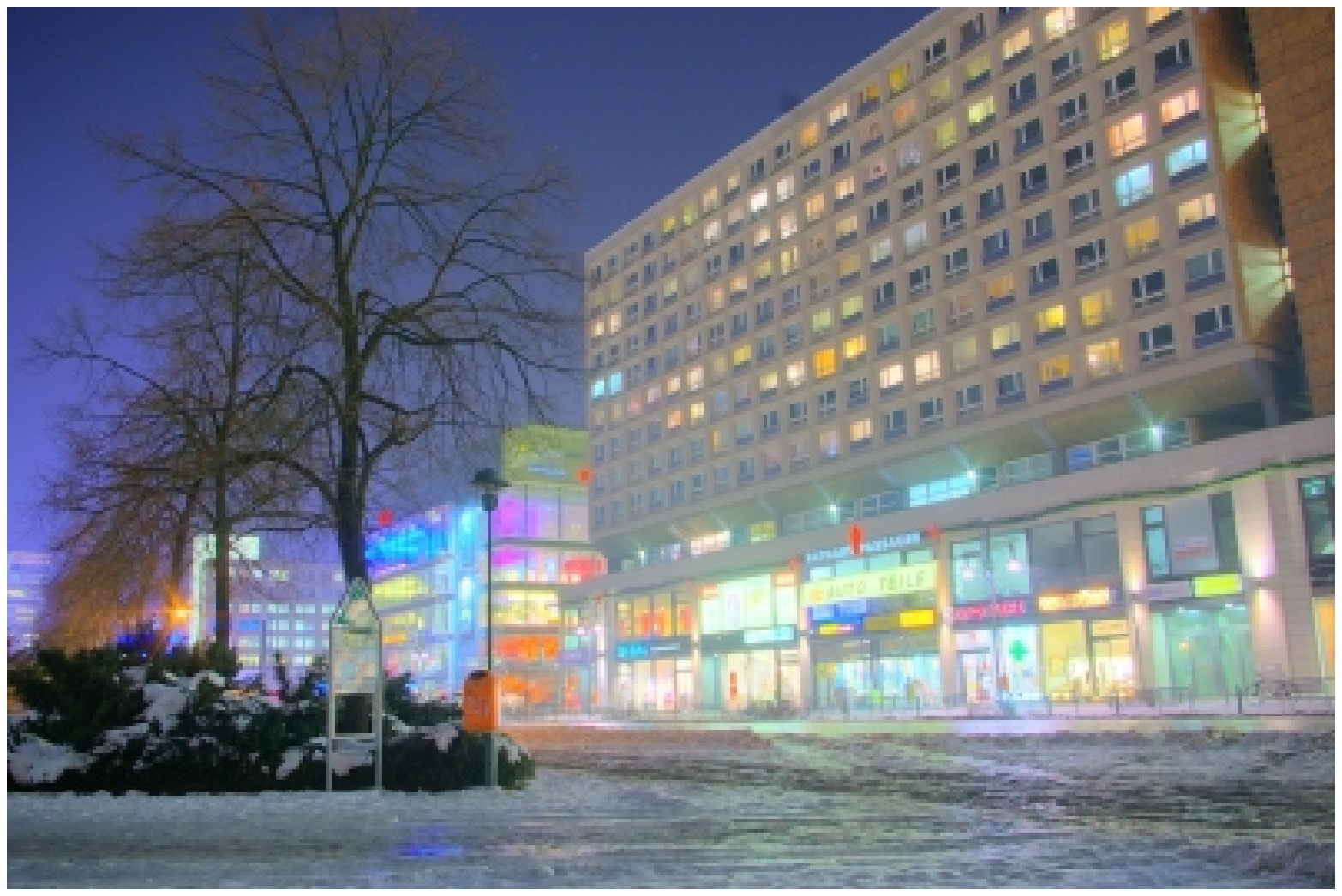} &
        \includegraphics[width=0.21\linewidth]{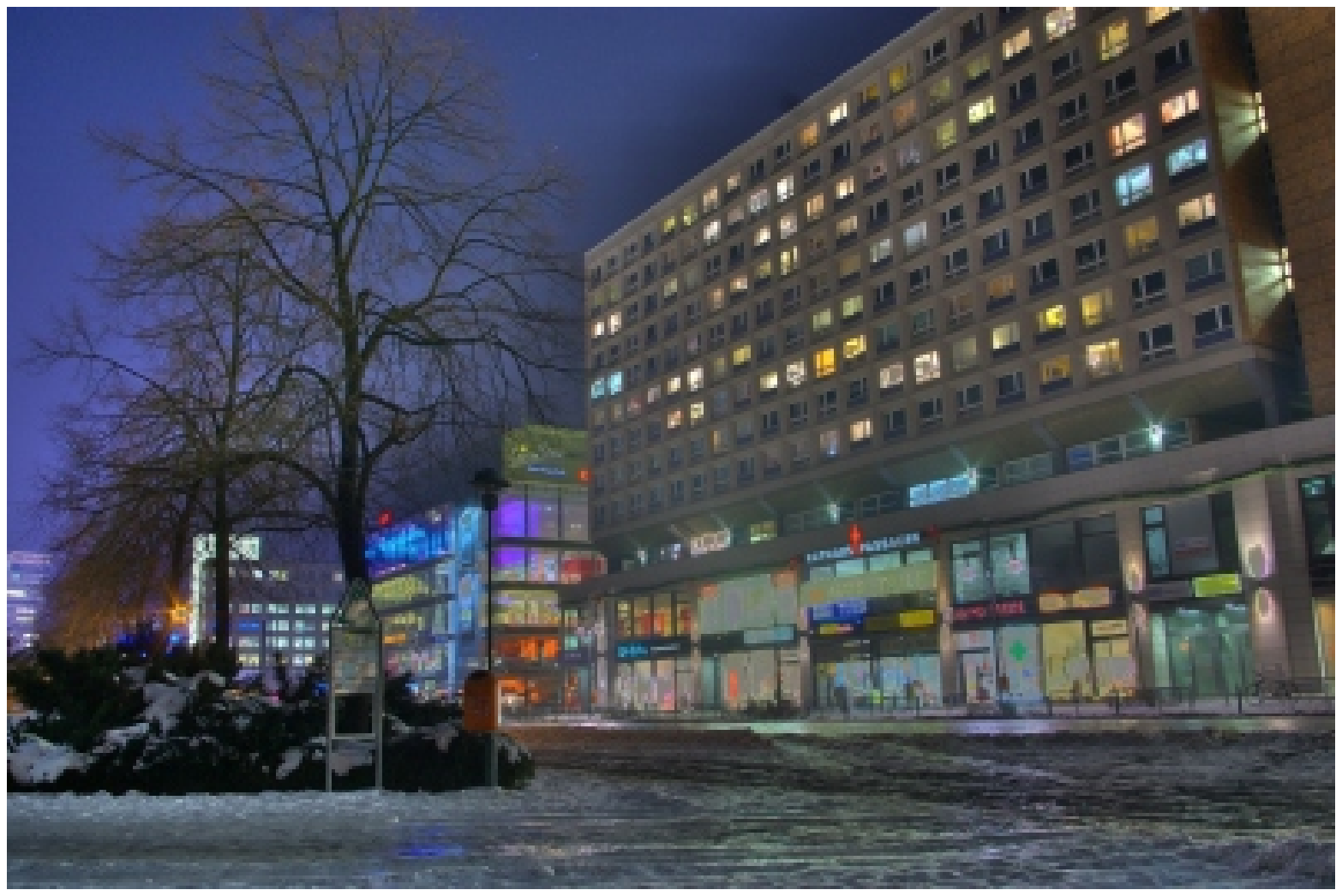} &
        \includegraphics[width=0.21\linewidth]{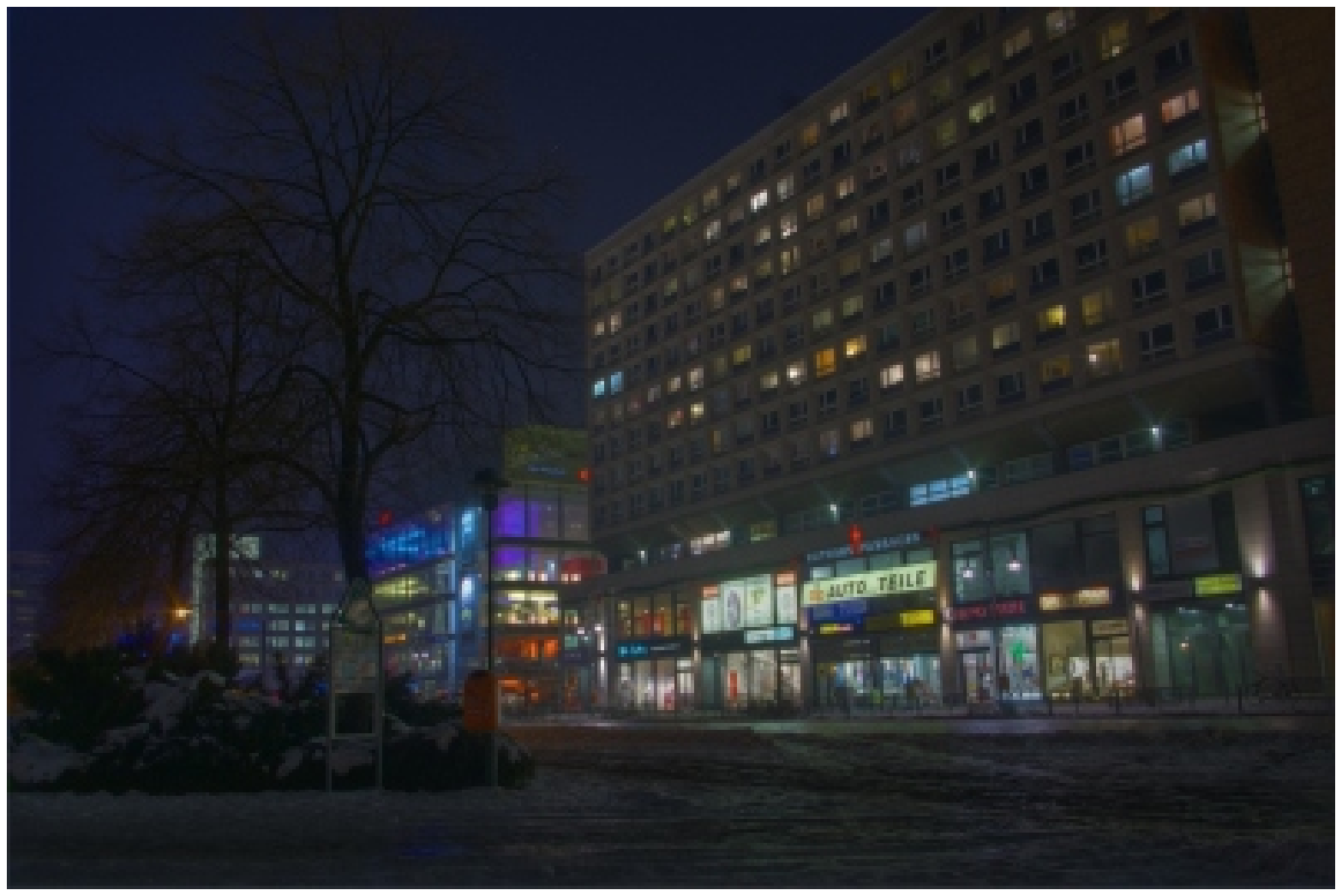} \\
            (e) Reinhard et al. \yrcite{Reinhard:2005} & (f) Drago et al. \yrcite{Drago:2003} &
            (g) Durand et al. \yrcite{Durand:2002} & (h) Fattal et al. \yrcite{Fattal:2002} \\
    \end{tabular}
    \caption{The resulting images obtained with  HDR state of the art imaging techniques (irradiance
    maps fusion followed by TMO and exposure fusion ). }
    \label{Fig:StateOfArt}
    \end{center}
\end{figure*}

For comparative results we considered the exposure fusion in the variant modified according to
\cite{Mertens:2007} and \cite{Zhang:2012} using the author released code and the TMOs applied on
the \texttt{.hdr} images described in \cite{Ward:97}, \cite{Fattal:2002}, \cite{Durand:2002},
\cite{Drago:2003}, \cite{Reinhard:2005}, \cite{Krawczyk:2005} and \cite{Banterle:2012} as they are
the foremost such methods. The code for the TMOs is taken from the Matlab HDR Toolbox
\cite{Banterle:2011} and is available online\footnote{The HDR toolbox may be retrieved from
\url{http://www.banterle.com/hdrbook/downloads/} $\phantom{XXXXX}$ \url{HDR_Toolbox_current.zip}}.
The implemented algorithms were optimized by the Toolbox creators to match with initial article
reported results and for better performance; hence, we used the implicit values for the algorithms
parameters. We note that envisaged TMO solutions include both global operators and local
adaptation. A set of examples with the results produced with all the methods is presented in Fig.
\ref{Fig:StateOfArt}.

\subsection{Objective metrics}

\begin{table}[t]
    \begin{center}
    \caption{HDR image evaluation by the average values for structural fidelity ($S$), statistical
    naturalness ($N$) and the overall quality ($Q$) and detailed
    in Section \ref{Sect:Implement}.1. With bold letters we marked the best result
    according to each category, while  with italic the second one. }
    \label{Tab:TQIM}
        \begin{tabular}{|c|c|c|c|}
        \hline
            \textbf{Method}            & $S$[\%] & $N$[\%]  & $Q$[\%]   \\ \hline \hline
            Ward et al. \yrcite{Ward:97}       & 66.9    & 14.38    & 72.7 \\ \hline
            Fattal et al. \yrcite{Fattal:2002}   & 59.9    & 6.4      & 61.0 \\ \hline
            Durand et al. \yrcite{Durand:2002}   & 81.7    & 41.0      & 85.4 \\ \hline
            Drago et al. \yrcite{Drago:2003}    & 82.3    & 50.2     & 87.0 \\ \hline
            Reinhard et al. \yrcite{Reinhard:2005} & \emph{83.1}    & 50.5     & 87.5 \\ \hline
            Krawczyk et al. \yrcite{Krawczyk:2005} & 71.7    & 36.8     & 76.6 \\ \hline
            Banterle et al. \yrcite{Banterle:2012} & \textbf{83.7}    & 52.1     & 87.8 \\ \hline \hline
           Mertens et al. \yrcite{Mertens:2007}  &  81.7    & \textbf{64.2}     & \textbf{89.4}  \\ \hline
           Zhang et al. \yrcite{Zhang:2012}    &  77.6    & \emph{59.7}     &  83.4  \\ \hline
           \hline
            \emph{Proposed}, $\mu=0.5$    & 81.0 & 39.2   & 84.5  \\ \hline
            \emph{Proposed}, $\mu=0.4$    & 81.6 & 52.1   & 87.3  \\ \hline
            \emph{Proposed}, $\mu=0.37$   & 81.5 & \emph{57.4}   & \emph{88.0}  \\ \hline
            \emph{Proposed}, $\mu=0.32$   & 81.4 & 53.7   & 87.4  \\ \hline
        \end{tabular}
    \end{center}
\end{table}

\emph{Structure and Naturalness. } We started the evaluation using the set of three objective
metrics from \cite{Yeganeh:2013}. The results obtained are presented in Table \ref{Tab:TQIM}. The
best performing version of the proposed method was for $\mu=0.37$.

The proposed method, when compared with various TMOs, ranked first, according to the overall
quality and statistical naturalness, and it rank fifth according to structural fidelity (after
\cite{Banterle:2012}, \cite{Reinhard:2005}, \cite{Drago:2003}, \cite{Durand:2002}).  Our method was
penalized when compared to other TMO due to their general adaptation being closer to the standard
contrast sensitivity function (CSF) \cite{Barten:1999}. Yet we stress that some TMOs
\cite{Krawczyk:2005} or \cite{Banterle:2012} work only for calibrated images in specific scene
luminance domain.

When compared with other exposure fusion methods \cite{Mertens:2007} and \cite{Zhang:2012}, it
ranked second for the overall quality after \cite{Mertens:2007}. This is an expected results as
standard exposure fusion was build to match subjective opinion score as the envisaged metrics did
too. Yet the proposed method outperformed the overall performance of the exposure fusion introduce
in \cite{Zhang:2012}. Furthermore a more recent algorithm, namely the ExpoBlend \cite{Bruce:2014}
reports the overall quality on two image that we used too (''Memorial'' and ''Lamp''). On these
images, the proposed method outperformed ExpoBlend: on ''Memorial'' we reach 95.5\% compared to
93.2\%, while on ''Lamp'' we reach 90.1\%  compared to 89.4\% reported in \cite{Bruce:2014}.

Furthermore, the proposed method is the closest to the standard exposure fusion result
\cite{Mertens:2007}, which is currently the state of the art method for consumer applications while
building HDR images. This aspect is shown in Table \ref{Tab:logRMSE}, where we computed the natural
logarithm of the Root-Mean-Square to the image resulting from the standard exposure fusion and,
respectively, the structural similarity when compared to the same image.

\begin{table}[t]
    \begin{center}
     \caption{HDR image evaluation by taking the log of root mean square to standard exposure fusion.
        Best values are marked with bold letters. }
    \label{Tab:logRMSE}
        \begin{tabular}{|c|c|c|}
        \hline
            \textbf{Method}            & logRMSE[dB]  & SSIM    \\ \hline \hline
            Ward et al. \yrcite{Ward:97}       & 149.9    & 83.5    \\ \hline
            Fattal et al. \yrcite{Fattal:2002}   & 281.1    & 38.1       \\ \hline
            Durand et al. \yrcite{Durand:2002}   & 160.5   & 74.4       \\ \hline
            Drago et al. \yrcite{Drago:2003}    & 148.4   & 74.7      \\ \hline
            Reinhard et al. \yrcite{Reinhard:2005} & 148.4    & 74.8      \\ \hline
            Krawczyk et al. \yrcite{Krawczyk:2005} & 136.9    & 66.4     \\ \hline
            Banterle et al. \yrcite{Banterle:2012} & 147.9    & 75.2      \\ \hline \hline
            \emph{Proposed}           & \textbf{72.1} & \textbf{93.8}    \\ \hline

        \end{tabular}
    \end{center}
\end{table}

\emph{Perceptualness.} One claim of the current paper is that the proposed method adds
perceptualness to the exposure fusion. To test this, we compared our method against the standard
exposure fusion, \cite{Mertens:2007} using the perceptual DRIM metric from \cite{Aydin:2008}. Over
the considered database, the proposed method produced an average total error (the sum of three
categories) with 2\% smaller than the standard exposure fusion (64.5\% compared to 66.8\%). On
individual categories, the proposed method produced a smaller amount of amplification of contrast,
with comparable results on loss and reversal of contrast. Thus, overall, the results confirm the
claim.

\subsection{Artifacts}
\begin{figure}[t]
    \begin{center}
    \begin{tabular}{cc}
        \includegraphics[width=0.43\linewidth]{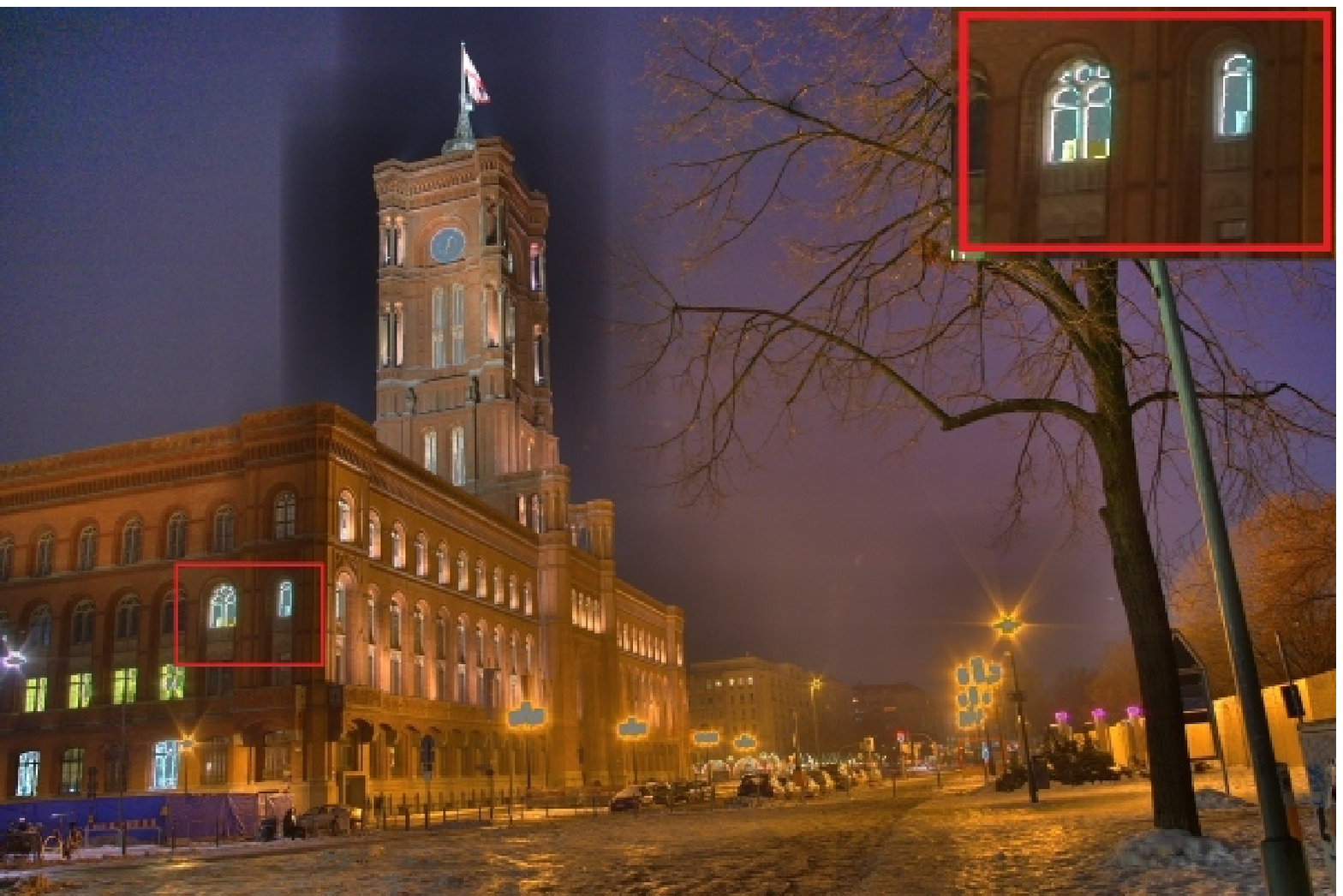} &
        \includegraphics[width=0.43\linewidth]{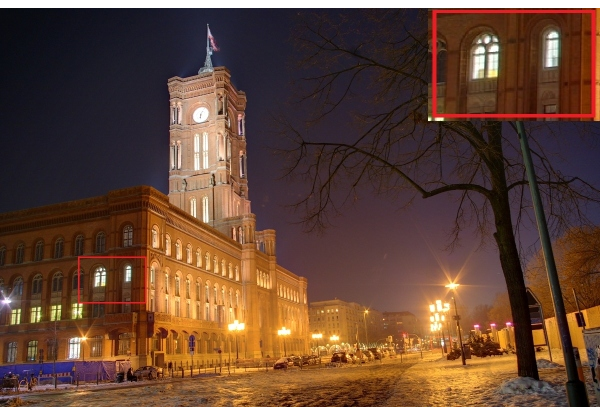} \\
        (a) Durand et al. \yrcite{Durand:2002} & (b) Proposed \\
    \end{tabular}
    \caption{Examples of artifacts produced by state of the art methods compared to the robustness
        of the proposed method. Close-ups point to artifact areas.}
    \label{Fig:Artifacts}
    \end{center}
\end{figure}

The HDR-specific objective metrics have the disadvantage of not properly weighting the artifacts
that appear in images, while human observers are very disturbed by them. This fact was also pointed
by \cite{Cadik:2008} and to compensate we performed visual inspection to identify disturbing
artifacts. The proposed method never produced any artifact in the tested image sets. Examples of
state of the art methods and artifacts produced may be seen in Fig. \ref{Fig:Artifacts}.

\begin{figure*}[t]
    \begin{center}
    \begin{tabular}{cc cc}
        \includegraphics[width=0.22\linewidth]{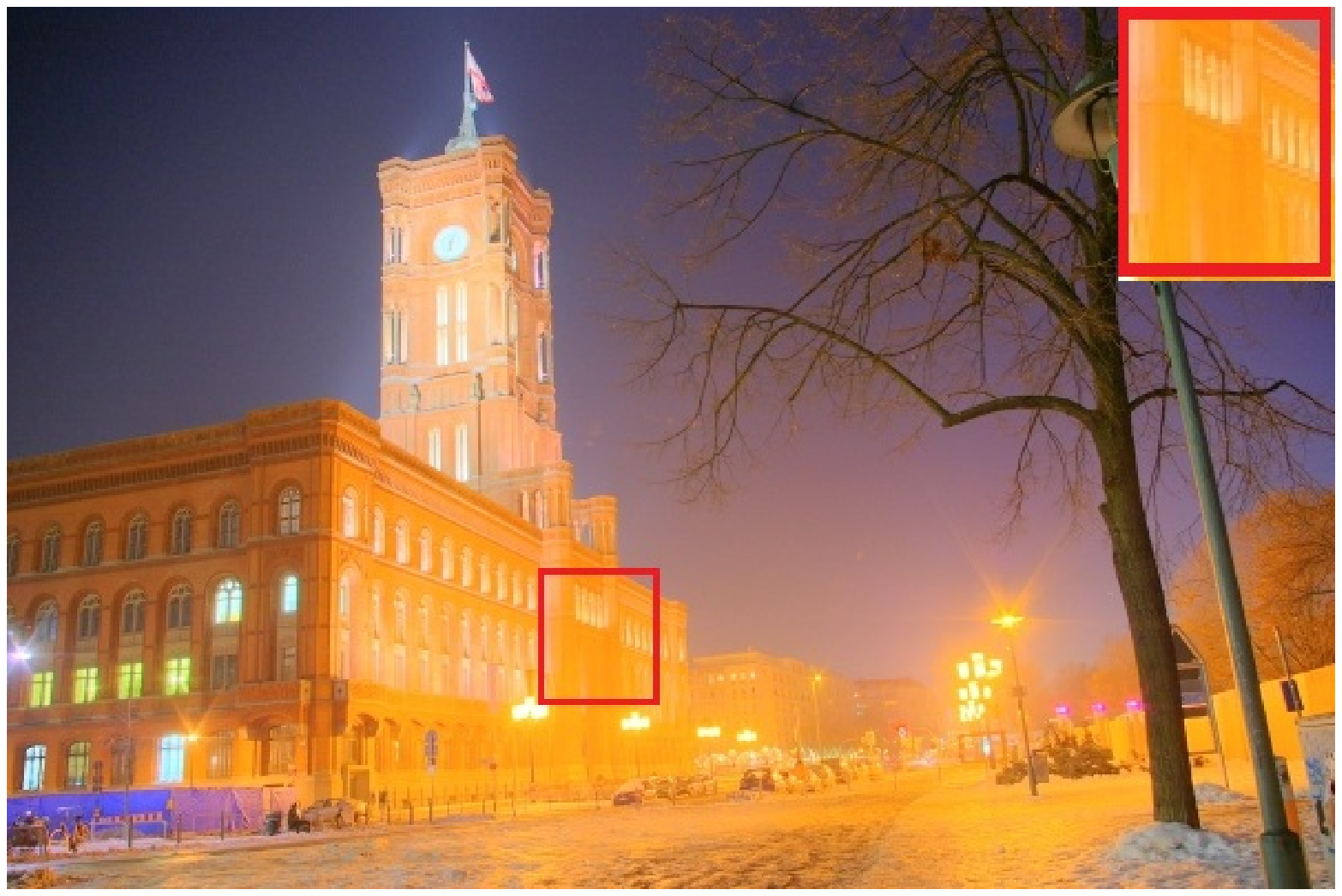} &
        \includegraphics[width=0.22\linewidth]{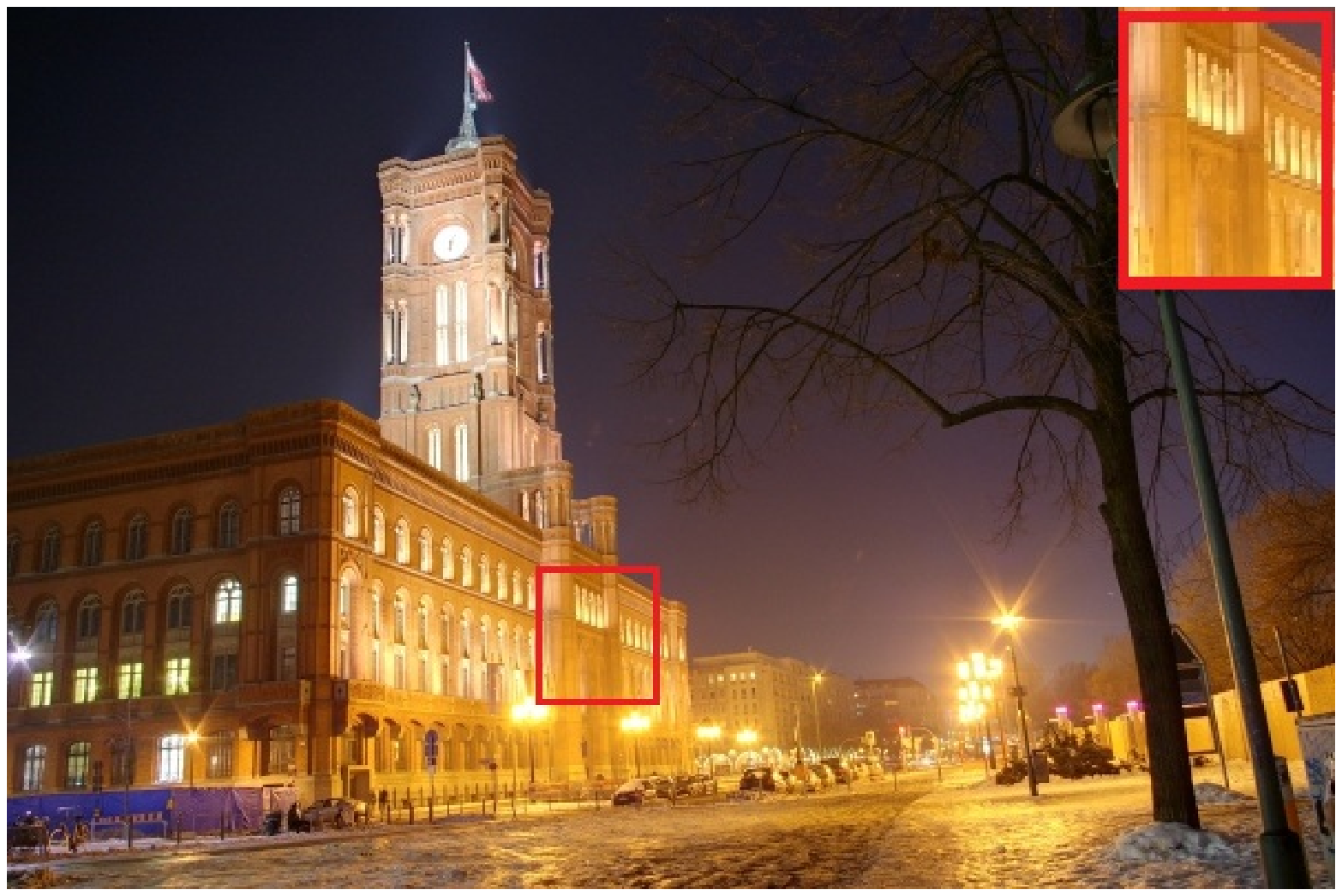} &
        \includegraphics[width=0.20\linewidth]{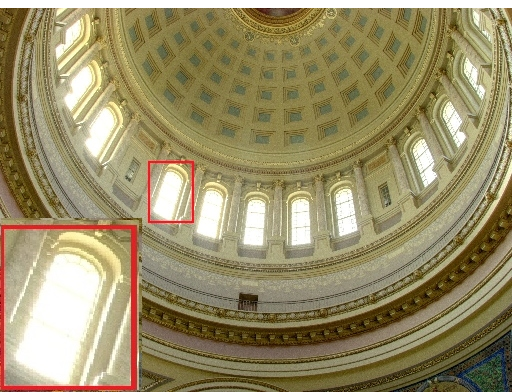} &
        \includegraphics[width=0.20\linewidth]{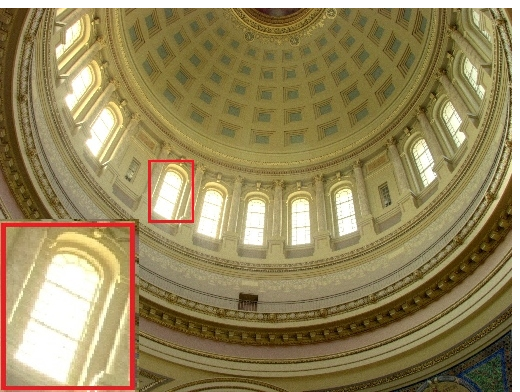} \\
        (a) Banterle et al. \yrcite{Banterle:2012} & (b) Proposed    &      (c) Mertens et al. \yrcite{Mertens:2007} & (d) Proposed\\
     \end{tabular}
    \caption{Examples of loss of details produced by state of the art methods compared to the
      robustness of the proposed method.}
    \label{Fig:LossOfDetail}
    \end{center}
\end{figure*}

In direct visual inspection, when compared against the standard exposure fusion method
\cite{Mertens:2007}, our algorithm shows details in bright areas, while normal, real--based
operations do not. This improvement is due to the closing property of the logarithmic addition and
respectively scalar amplification. This aspect is also visible when comparing with the most robust
TMO based method, namely \cite{Banterle:2012}. Examples that illustrate these facts are presented
in Fig. \ref{Fig:LossOfDetail}.

\subsection{Subjective ranking}
The non-experts ranked the images produced with the proposed method, standard exposure fusion, the
methods from \cite{Banterle:2012}, \cite{Pece:2010}, \cite{Drago:2003} and \cite{Reinhard:2005}.
Regarding the results, the proposed method was selected as the best one by 10 users, the exposure
fusion \cite{Mertens:2007} by 7, while the rest won 1 case. Also the second place was monopolized
by the "glossier" exposure fusion based method.

When compared to direct exposure fusion proposed by Mertens et al. \yrcite{Mertens:2007}, due to
the perceptual nature of the proposed method, a  higher percentage of the scene dynamic range is in
the visible domain; direct fusion losses information in the dark-tones domain and respectively in
very bright part; this is in fact the explanation for the narrow margin of our method advance.

\section{Discussion and Conclusions}
In this paper we showed that the LTIP model is compatible with the Naka-Rushton equation modelling
the light absorption in the human eye and similar with the CRF of digital cameras. Upon these
findings, we asserted that it is possible to treat different approaches to HDR imaging unitary.
Implementation of the weighted sum of input frames is both characteristic to irradiance map fusion
and to exposure fusion. If implemented using LTIP operations, the perceptualness is added to the
more popular exposure fusion. Finally, we introduced a new HDR imaging technique that adapts the
standard exposure fusion to the logarithmic type operations, leading to an algorithm which is
consistent in both theoretical and practical aspects. The closing property of the LTIP operations
ensures that details are visible even in areas with high luminosity, as previously shown.

The method maintains the simplicity of implementation typical to the exposure fusion, since the
principal difference is the redefinition of the standard operations and different parameter values.
The supplemental calculus associated with the non-linearity of LTIP operation could easily be
trimmed out by the use of look-up tables, as shown in \cite{Florea:2013}.

The evaluation results re-affirmed that in an objective assessment aiming at naturalness and
pleasantness of the image, the proposed method outperforms irradiance map fusion followed by TMOs
as they try to mimic more a theoretical model which is not perfect and is not how the normal user
expects HDR images to look. The same conclusion was emphasized by the subjective evaluation, where
methods developed in the image domain are preferred as the resulting images are more "appealing".
The method outperformed, even by a small margin, the standard exposure fusion when evaluated with
DRIM metric showing that is more HVS oriented. The proposed method, having a HVS inspired global
adaptation and "glossy" tuned local adaptation, by a narrow margin, ranks best in the subjective
evaluation.

\section*{Acknowledgments}
The authors wish to thank Martin Cadik for providing the frames for testing. We also would like to
thank Tudor Iorga for running tests and providing valuable ideas. The work has been partially
funded by the Sectoral Operational Programme Human Resources Development 2007-2013 of the Ministry
of European Funds through the Financial Agreement POSDRU/159/1.5/S/134398.

\bibliographystyle{dcu}
\bibliography{PerceptExpMerging}

\section{Biographies}
{
    \textbf{Corneliu Florea}
    born in 1980, got his master degree from University ``Politehnica'' of Bucharest
    in 2004 and the PhD from the same university in 2009. There, he lectures on statistical signal and
    image processing, and has introductory courses in computational photography and computer vision. His
    research interests include non-linear image processing algorithms for digital still cameras and
    computer vision methods for portrait understanding. He is author of more than 30 peer-reviewed
    papers and of 20 US patents and patent applications.

    \textbf{Constantin Vertan}
    Professor Constantin Vertan holds an image processing and analysis tenure at the Image Processing
    and Analysis Laboratory (LAPI)
    at the Politehnica University of Bucharest. 
    For his contributions in image processing he was awarded with
    UPB ''In tempore opportuno'' award (2002), Romanian Research Council ''In hoc signo vinces'' award
    (2004) and was promoted as IEEE Senior Member (2008). His research interests are: general image
    processing and analysis, content-based image retrieval, fuzzy and medical image processing
    applications. He authored more than 50 peer-reviewed papers. He is the secretary of the Romanian
    IEEE Signal Processing Chapter and associate editor at EURASIP Journal on Image and Video
    Processing.

    \textbf{Laura Florea}
    received her PhD in 2009 and M.Sc in 2004 from University ``Politehnica'' of Bucharest. From 2004
    she teaches classes in the same university, where she is currently a Lecturer. Her interests
    include image processing algorithms for digital still cameras, automatic understanding of human
    behavior by analysis of portrait images, medical image processing and statistic signal processing
    theory. Previously she worked on computer aid diagnosis for hip joint replacement. She is author of
    more than 30 peer-reviewed journal and conference papers.

}

\end{document}